\PassOptionsToPackage{table,dvipsnames}{xcolor}

\documentclass{maskwam}

\usepackage{amsmath}
\usepackage{amssymb}
\usepackage{amsfonts}
\usepackage{mathtools}
\usepackage{enumitem}
\usepackage{algorithm}
\usepackage{algorithmic}
\usepackage{wrapfig}
\usepackage{adjustbox}
\usepackage{tabularx}
\usepackage{makecell}
\usepackage{siunitx}
\usepackage{float}
\usepackage{diagbox}
\usepackage{dsfont}

\definecolor{bestgreen}{RGB}{224,232,240}
\definecolor{secondgreen}{RGB}{240,245,249}

\usepackage[utf8]{inputenc} %
\usepackage[T1]{fontenc}    %
\usepackage{url}            %
\usepackage{booktabs}       %
\usepackage{amsfonts}       %
\usepackage{nicefrac}       %
\usepackage{microtype}      %
\usepackage[table,dvipsnames,rgb]{xcolor}
\usepackage{makecell}
\usepackage{bm}
\usepackage{amsmath}
\usepackage{graphicx}
\usepackage{wrapfig}
\usepackage{multirow}
\usepackage{subcaption}
\usepackage{enumitem}
\usepackage{colortbl}

\newcommand{\question}{%
    \stepcounter{question}%
    \noindent\textbf{Q\thequestion:~\ignorespaces}%
}

\newcounter{question}
\definecolor{baselinecolor}{gray}{.9}

\newcolumntype{x}[1]{>{\centering\arraybackslash}p{#1pt}}
\definecolor{drp-blue}{HTML}{1f77b4}
\definecolor{pretty-blue}{RGB}{0, 113, 188}
\definecolor{kaiming-green}{RGB}{57,181,74} %
\definecolor{mypurple}{RGB}{55,0,168} %
\definecolor{icmlblue}{rgb}{0,0.08,0.45} %
\definecolor{mygreen}{HTML}{4FC978}
\definecolor{linecolor1}{RGB}{246, 248, 239}
\definecolor{linecolor2}{RGB}{230, 234, 217}
\definecolor{linecolor3}{RGB}{211, 222, 190}
\definecolor{reconcolor}{HTML}{412F8A}
\definecolor{runpei-orange}{HTML}{F35F27}
\definecolor{runpei_blue}{HTML}{14294B}
\definecolor{datacolor}{HTML}{0009BF}
\definecolor{vitcolor}{HTML}{fc8e62}
\definecolor{cvprblue}{rgb}{0.21,0.49,0.74}
\definecolor{myblue}{rgb}{.39,.58,.93}
\newcommand{\method}{ImageWAM}

\newcommand{\cmark}{\ding{51}}%
\newcommand{\xmark}{\ding{55}}%

\usepackage{pifont}

\usepackage{graphicx}

\usepackage{amssymb} 
\usepackage{marvosym}

\title{ImageWAM: Do World Action Models Really Need Video Generation, or Just Image Editing?}

\renewcommand{\authorlist}{%
  {\sffamily \sbseries
  Yuyang Zhang   $^{123*}$,
  Wenyao Zhang   $^{123*\dagger}$,
  Zekun Qi       $^{4}$,
  He Zhang       $^{3}$,
  Haitao Lin     $^{3}$,
  Jingbo Zhang   $^{3}$,\\
  Yao Mu         $^{1}$,
  Xiaokang Yang  $^{1}$,
  Wenjun Zeng    $^{2}$,
  Xin Jin        $^{25\text{\Letter}}$
  }%
}

\affiliation[1]{Shanghai Jiao Tong University}
\affiliation[2]{Eastern Institute of Technology}
\affiliation[3]{Tencent Robotics X}
\affiliation[4]{Tsinghua University}
\affiliation[5]{Zhongguancun Academy}

\contribution[*]{Equal contribution}
\contribution[\dagger]{Project Lead}
\contribution[\text{\Letter}]{Corresponding author}

\abstract{
World Action Models (WAMs) commonly rely on video generation to bridge visual world modeling and robot control. 
However, video-based WAMs face three coupled limitations: dense multi-frame future tokens make inference costly, full video prediction spends capacity on action-irrelevant temporal and appearance details, and long-horizon future imagination may introduce errors that mislead action prediction. 
These issues raise a simple question: \textit{Does world action model really need video generation}?
We propose \textbf{\method{}}, a simple WAM framework that repurposes pretrained image editing models for robot action prediction. 
In contrast to video generation, image editing provides a better-matched prior: it only needs to model a target-frame transformation, focuses on action-relevant current-to-target visual differences, and grounds task instructions to localized visual changes through edit pretraining. 
In practice, \method{} does not decode the target frame at inference time; instead, it conditions a flow-matching action expert on the KV caches produced by image-editing denoising, using them as a compact world-action context.
\method{} outperforms standard VLA baselines and matching competitive WAMs without additional policy pretraining across different simulator and real-world experiments. 
It also reduces FLOPs to 1/6 and latency to 1/4 of video-based WAMs.
Attention analysis further shows that editing caches focus on task-relevant change regions, supporting image editing as an effective alternative to video-based world-action modeling.
}

\date{\sffamily\today}

\metadata[Project Page]{\url{https://zhangwenyao1.github.io/ImageWAM/}}
\metadata[Github]{\url{https://github.com/yuyangalin/ImageWAM}}

\begin{document}
\maketitle
\vspace{-3mm}
\begin{figure}[h!]
    \centering
    \includegraphics[width=0.99\linewidth]{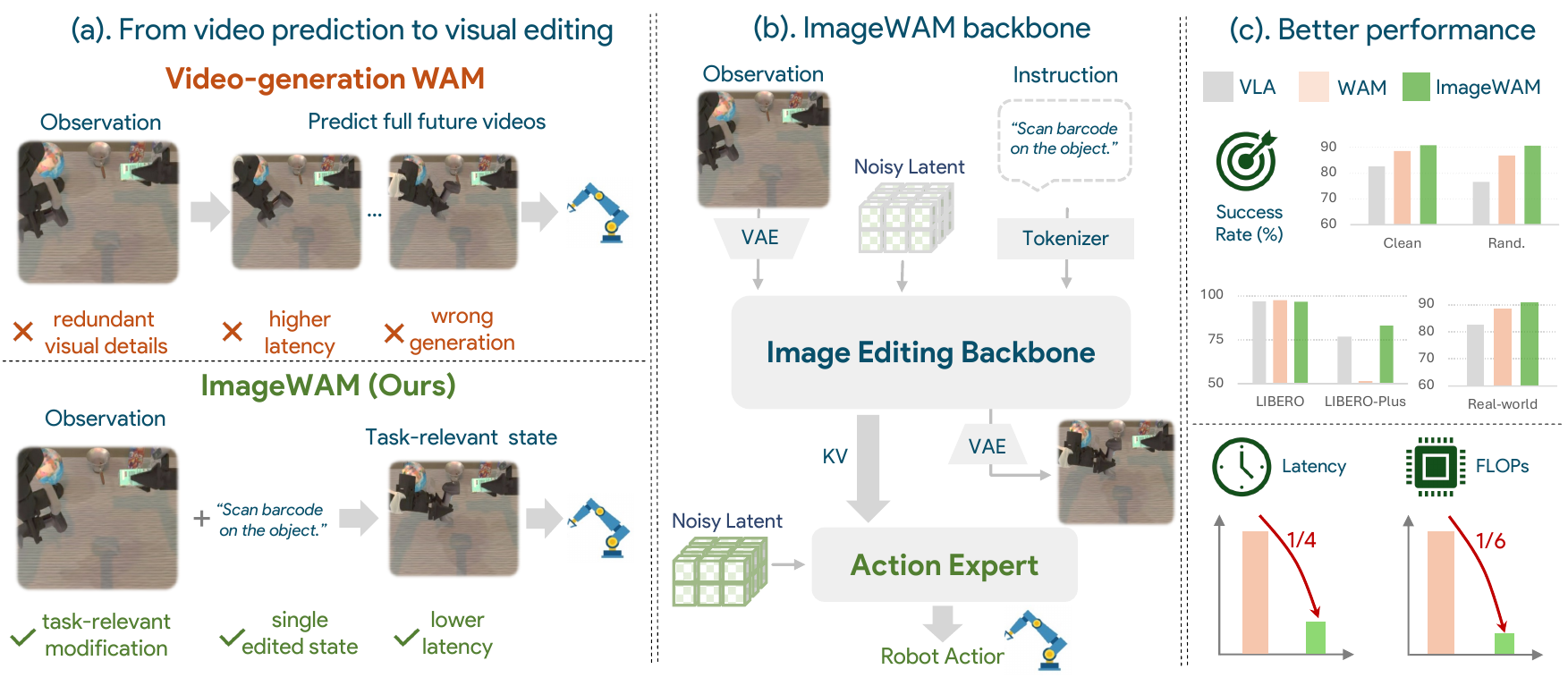}
    \vspace{-3mm}
    \caption{Previous video-generation WAMs instantiate world-action reasoning by predicting dense future video tokens, which can be computationally expensive and may allocate capacity to action-irrelevant visual details.
\method{} replaces future video prediction with an image editing backbone that reasons over a source-grounded, instruction-guided visual transformation.
The resulting edit-aware representation serves as a compact world-action intermediate for action prediction, achieving strong policy performance while reducing inference cost.}
    \label{fig:teaser}
\end{figure}
\vspace{-1mm}

\section{Introduction}

Recent robot policy learning has increasingly explored video generation models as world-action backbones. 
This direction is appealing because video pretraining exposes models to rich visual dynamics, such as object motion, temporal continuity, physical interaction, and scene evolution~\cite{vpp2024hu,ye2026dreamzero,li2026lingbotva,madit4dit,kim2026cosmos}. 
It also supports a reason-before-act paradigm: a policy may first imagine how the scene will change, and then use this imagined future to guide action prediction~\cite{hu2026bagelvla,zhang2026uam,fan2026aim}. 
Together with the scalability of generative pretraining on large and heterogeneous video data~\cite{zhu2025unified,lyu2026lda,zhang2026disentangled,bi2025motus}, video models provide an intuitive bridge between visual world modeling and robot control.

However, this bridge also reveals a mismatch as shown in Figure~\ref{fig:teaser}(a). 
Video generation models are trained to synthesize complete future videos. 
To do so, they must model appearance details, background changes, camera motion, temporal smoothness, and many other factors that may be only weakly related to the robot's next action~\cite{yuan2026fastwam,ye2026gigaworldpolicy,yu2026maskwam}.
Generating many spatio-temporal tokens across multiple frames makes inference costly for real-time robot control~\cite{ye2026dreamzero,li2026lingbotva}.
Moreover, generating a physically consistent video is a hard proxy task~\cite{peng2026reworld,yang2025orv,zhen2025tesseract}. 
This is especially true for fine-grained manipulation, where small contact events, slight object displacements, or subtle configuration changes can determine success, but are difficult to predict reliably over multiple frames. 
If the imagined video is wrong, the downstream action predictor may be misled. 
These issues raise a simple question: \textit{Does the world action model really require video generation?}

We argue that image editing models offer a more direct visual generative prior for language-conditioned manipulation.
Instead of predicting how an entire scene evolves over time, image editing models are trained to transform a source image according to a language instruction.
This objective matches a key requirement of robot policies: the model should understand what task-relevant visual change should happen in the current scene under the given instruction.
For many manipulation tasks, the essential signal is not a photorealistic future video, but an instruction-guided transformation from the current observation toward a desired visual state as illustrated in Figure~\ref{fig:teaser}(a).

This view gives image editing models three advantages as robot policy backbones. First, they provide strong instruction-to-change alignment. Their pretraining objective directly couples language with visual modifications, encouraging the model to focus on what should change, where it should change, and how the change is specified by the instruction. 
Second, editing provides an easier and more action-relevant proxy than full video prediction. Rather than modeling complete temporal trajectories, an editing model focuses on the visual difference between the current state and an instruction-consistent target state. 
This avoids spending capacity on irrelevant temporal details and reduces the risk of using inaccurate future videos for action generation. 
Third, editing offers a more compact inference path. A policy can use internal editing-aware representations that encode the intended visual transformation, without decoding dense multi-frame videos at inference time.

Motivated by this insight, we propose \textbf{\method{}}, a new framework that repurposes pretrained image editing models as backbones for robot action prediction, as shown in Figure~\ref{fig:teaser}(b). 
Given the current observation and task instruction, \method{} extracts editing-aware representations from an image editing backbone and feeds them into an action prediction head. 
Our goal is not to generate visually appealing edited images, nor to use editing models as goal-image generators. 
Instead, we use their intermediate instruction-conditioned features as transformation-aware representations for direct policy learning. 
This design preserves the benefits of generative visual pretraining while avoiding explicit future video synthesis, leading to a compact inference path for real-time control.

Empirically, we find that editing-aware representations are effective for language-conditioned robot policies. 
Under comparable action prediction architectures, \method{} improves over standard visual and vision-language backbones, showing that the gains are not merely due to stronger image recognition or language alignment. 
Our analyses further show that instruction conditioning and editing-oriented feature extraction are important for obtaining action-relevant representations. 
These results suggest that image editing models provide a promising backbone choice for robot policy learning, broadening visual generative pretraining beyond video-based world modeling.

Our contributions are three-fold:

\begin{itemize}[leftmargin=1.5em,topsep=1.5pt,itemsep=1.5pt]
\item We introduce \method{}, a framework that repurposes pretrained image editing models as instruction-conditioned visual backbones for robot action prediction, offering an alternative to video-generation-based world action models.
\item We formulate robot manipulation as instruction-guided visual transformation and identify three properties of image editing pretraining that are well aligned with policy learning: instruction-to-change alignment, easier goal/change proxy, and compact inference.
\item We empirically validate the effectiveness of editing-aware representations against standard visual and vision-language backbones, and analyze the role of instruction conditioning and editing-oriented feature extraction in action prediction.

\end{itemize}

\section{Related Works}
\subsection{Image Editing}
Text-guided image editing modifies a source image according to a language instruction while preserving irrelevant content~\cite{wang2024scene,nanobananapro2025,gpt1p5,ye2025imgedit,wu2025qwen,glm_image,nextstep,longcatnext,zheng2026uni,team2025zimage}. 
Recent diffusion-based and MLLM-enhanced editing models have progressed from simple object-level edits to more complex spatial, semantic, and knowledge-driven modifications~\cite{zhang2023magicbrush,fu2024mgie,sheynin2024emuedit,yu2025anyedit,gabeur2026image,wang2026diffusion,jeanson2026leveraging}. 
While prior work mainly focuses on perceptual quality and instruction fidelity, we study image editing from a robotics perspective, using its source-conditioned and change-centric representations as compact world-action backbones for robot policy learning.

\subsection{World Action Models}

Unlike vision language action models~\cite{24pi0,25pi0.5,bjorck2025gr00t,zhang2025dreamvla,song2026reconvla,team2026hy,lin2026universal,yuan2025depthvla,qu25spatialvla,seer24,cotvla25,zhang2025digflow,beingbeyond2025beingh0,chen2025unified,li2025spatial,sun2026vla,wang2026vla,wu2026pragmatic,lee2025molmoact,Lv2025F1AV,wang2026qwen,xu2025seeing}, video generation models have recently been explored as predictive priors for robot policy learning. 
Early world action model~\cite{du2024learning,susie23,feng2025vidar,wen2024vidman}
 treats video generation as an explicit visual planning model: given the current observation and task context, the model predicts a complete future video or visual rollout, which is then translated into executable actions by an inverse dynamics model or action decoder~\cite{Liu2026WorldAV,Chen2025LargeVP,tan2025anypos,Mi2026TCIDMGV,Zhang2026VeoActHF,ge2026vampo,zhang2026world}. 
More recent works broaden this paradigm by using video generative models as representation extractors for action generation~\cite{li2025worldeval,xu2026kinema4d,jiang2026wovr,zhang2026robostereo,kim2026cosmos,chen2026unit,bardhan2026persistent,wei2026fate,lang2026vag,wang2026interactive,liu2026world}, value prediction~\cite{li2026world} and interactive world modeling~\cite{Liao2025GenieEA,Li2026dWorldEvalSR,Wang2026InteractiveWS,agarwal2026cosmos}. 
However, they are still largely built around video generation priors. 
Such designs often require predicting or processing dense spatio-temporal future tokens, leading to non-trivial inference cost and potentially modeling action-irrelevant and unrealistic visual details.
\method{} uses instruction-guided editing caches as a compact world-action context, avoiding dense future-video token processing while preserving the advantage of WAMs.

\section{Method}

\subsection{Problem Formulation}

We consider robot manipulation conditioned on a current visual observation and a task instruction. 
At each time step $t$, the robot receives an image observation $o_t$ and a task instruction $l$, and predicts an action chunk
\begin{equation}
    \mathbf{a}_{t:t+H} = (a_t, a_{t+1}, \ldots, a_{t+H}),
\end{equation}
where $H$ denotes the action horizon. 
The policy learning objective is
\begin{equation}
    \pi_\theta(\mathbf{a}_{t:t+H} \mid o_t, l).
\end{equation}

World-action models introduce an intermediate visual reasoning step before action prediction. 
Video-generation-based WAMs typically instantiate this intermediate by predicting a future visual trajectory:
\begin{equation}
    (o_t, l) \rightarrow \hat{o}_{t+1:t+H+1} \rightarrow \mathbf{a}_{t:t+H}.
\end{equation}
This enables reason-before-act policy learning, but requires generating dense spatio-temporal visual tokens across multiple future frames. 
Instead of predicting the full future trajectory, Our \method{} predicts only the endpoint frame:
\begin{equation}
    (o_t, l) \rightarrow \hat{o}_{\mathrm{edit}} \equiv \hat{o}_{t+H+1} \rightarrow \mathbf{a}_{t:t+H}.
\end{equation}
$\hat{o}_{\mathrm{edit}}$ is a single source-conditioned frame that summarizes the task-specified visual transformation of the current observation. 
It serves as a compact world-action intermediate for action prediction.

\begin{figure}[t]
    \centering
    \includegraphics[width=0.9\linewidth]{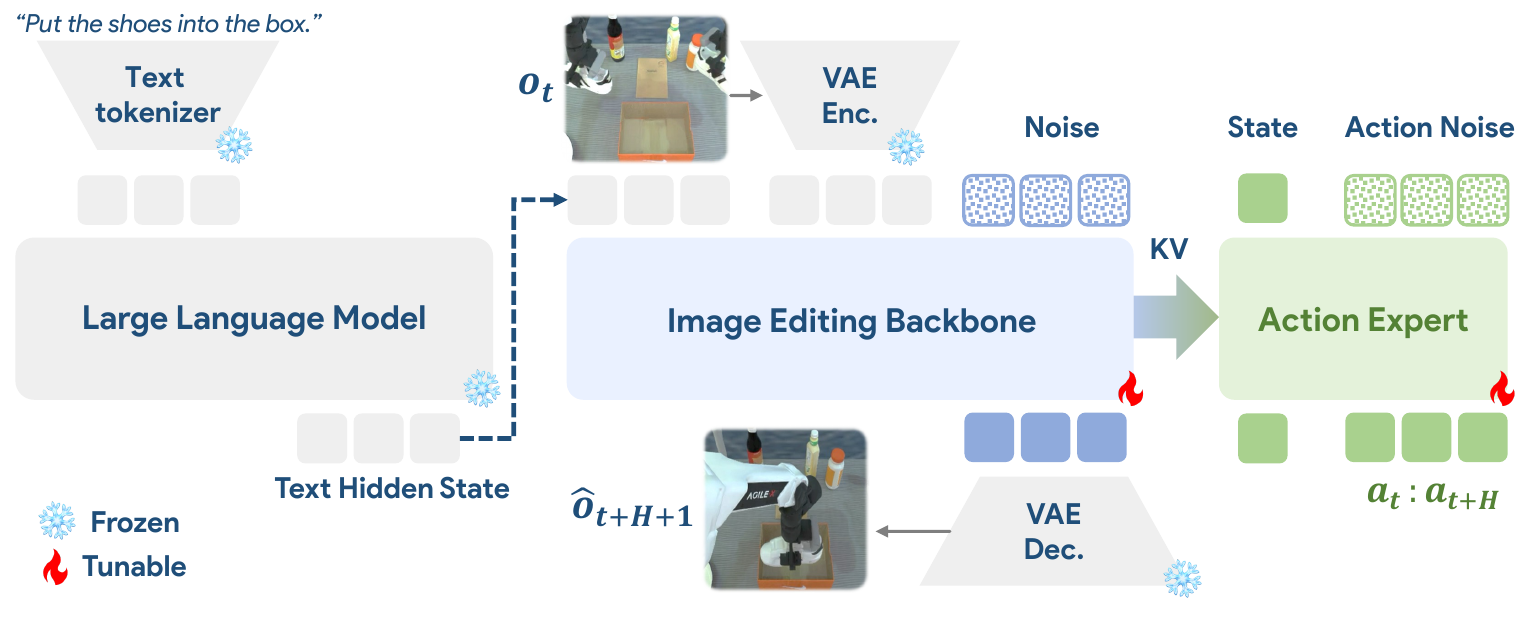}
    \caption{\textbf{ImageWAM Pipeline.} Given a language instruction and the current observation $o_t$, the image editing backbone synthesizes the future frame $\hat{o}_{t+H+1}$. The Action Expert integrates the intermediate KV features from this generation process via joint attention, predicting a sequence of future actions $\mathbf{a}_{t:t+H}$ conditioned on the current robot state and action noise.}
    \label{fig:imagewam_pipeline}
    \vspace{-2mm}
\end{figure}

\subsection{\method{} Architecture}

\method{} builds on a variant image editing model like OmniGen2~\cite{wu2025omnigen2},Ovis-U1~\cite{wang2025ovisu1} and Flux2~\cite{flux-2-2025} by attaching an action expert to their image editing branch. 
OmniGen2 provides a source-conditioned image editing backbone that takes the current observation $o_t$ and task instruction $l$ as inputs. 
Instead of using the editing branch only to decode an edited image, \method{} reuses the intermediate transformer key-value caches produced during denoising as conditioning context for action generation.

During training, we randomly sample an editing denoising timestep $\tau$ and run the editing branch at this timestep. 
For each transformer layer $\ell$, we collect the corresponding key-value cache:
\begin{equation}
    \mathcal{C}_{\mathrm{edit}}^{\tau}
    =
    \left\{
    (K_{\ell}^{\tau}, V_{\ell}^{\tau})
    \right\}_{\ell=1}^{L}
    =
    f_{\mathrm{edit}}^{\tau}(o_t, l),
\end{equation}
where $L$ is the number of transformer layers. 
The cache $\mathcal{C}_{\mathrm{edit}}^{\tau}$ is computed after the visual latent has interacted with the task instruction through the editing backbone. 
It therefore contains task-conditioned visual transformation information without requiring the final edited image to be decoded.

The action expert conditions on $\mathcal{C}_{\mathrm{edit}}^{\tau}$ for action generation. 
This design transfers the image editing model's internal reasoning process to robot control: the editing branch reasons about how the source observation should change under the task instruction, while the action expert converts this editing context into executable robot actions. 
Unlike video-generation WAMs, \method{} does not require future video tokens to be generated or decoded.

In addition to the standard video-WAM variant that performs denoising over future video tokens, we also implement a Fast-WAM-style variant~\cite{yuan2026fastwam}. 
In this variant, future video tokens are used only during training for video co-training, but are removed at inference time. 
The action expert is conditioned on the KV caches produced from the current observation and task instruction, without instantiating or denoising future video tokens. This gives a video-WAM baseline with the same no-future-token inference interface as Fast-WAM.

We keep the VLM and multimodal understanding components of the editing model frozen, including the modules used to encode task instructions and visual context. 
Only the diffusion-based image generation branch and the action expert are updated during training. 
The frozen VLM provides stable language-vision conditioning, while the trainable diffusion branch learns to predict task-relevant future frames and to produce editing caches useful for action generation.

\subsection{Action Prediction and Training}

\paragraph{Image editing objective.}
The editing branch is trained to predict a task-relevant future endpoint frame. 
Let $o_{t+H+1}$ denote the target future observation and let $z^{*}_{t+H+1}=E_{\mathrm{vae}}(o_{t+H+1})$ be its latent representation. 
We sample image noise $\epsilon_z \sim \mathcal{N}(0,I)$ and an image flow time $r \in (0,1)$, and construct the interpolated image latent
\begin{equation}
    z_r
    =
    (1-r) z^{*}_{t+H+1s}
    +
    r \epsilon_z .
\end{equation}
The diffusion image branch predicts the corresponding velocity field:
\begin{equation}
    \mathcal{L}_{\mathrm{img}}
    =
    \mathbb{E}_{z^{*}, \epsilon_z, r}
    \left[
    \left\|
    u_\phi(
        z_r,
        r
        \mid
        o_t, l
    )
    -
    (\epsilon_z - z^{*}_{t+K})
    \right\|_2^2
    \right],
\end{equation}
where $u_\phi$ denotes the velocity predictor of the diffusion image branch. 
This objective preserves the editing branch's ability to predict task-relevant future visual states and encourages the extracted editing caches to encode useful visual transformation information.

\paragraph{Action flow matching.}
The action expert generates an action chunk using a flow-matching objective. 
Let $\mathbf{a}^{*}_{t:t+H}$ denote the expert action chunk and let $\epsilon_a \sim \mathcal{N}(0,I)$ be Gaussian noise. 
We sample an action flow time $s \in (0,1)$ and construct the interpolated action sample
\begin{equation}
    \mathbf{a}_s
    =
    (1-s)\mathbf{a}^{*}_{t:t+H}
    +
    s\epsilon_a .
\end{equation}
Conditioned on the current observation, task instruction, and editing context cache $\mathcal{C}_{\mathrm{edit}}^{\tau}$, the action expert predicts the velocity field:
\begin{equation}
    \mathcal{L}_{\mathrm{act}}
    =
    \mathbb{E}_{\mathbf{a}^{*}, \epsilon_a, s, \tau}
    \left[
    \left\|
    v_\theta(
        \mathbf{a}_s,
        s
        \mid
        o_t, l, \mathcal{C}_{\mathrm{edit}}^{\tau}
    )
    -
    (\epsilon_a - \mathbf{a}^{*}_{t:t+H})
    \right\|_2^2
    \right].
\end{equation}
Here, $s$ denotes the action flow-matching time, while $\tau$ denotes the image editing denoising timestep used to extract the editing cache. 
Sampling $\tau$ during training exposes the action expert to editing caches from different stages of the denoising process.
We jointly optimize the diffusion image branch and the action expert with $\mathcal{L}=\mathcal{L}_{\mathrm{act}}+\mathcal{L}_{\mathrm{img}}$.

\subsection{Efficient Inference}

At inference time, \method{} avoids full future-video generation and also does not require decoding a complete edited image. 
Instead of running the full image editing denoising trajectory, we select a fixed editing denoising timestep $\tau^\star$ and perform only one editing-branch forward step to obtain
\begin{equation}
    \mathcal{C}_{\mathrm{edit}}^{\tau^\star}
    =
    f_{\mathrm{edit}}^{\tau^\star}(o_t, l).
\end{equation}
Action expert generates the action chunk by denoising action samples conditioned on this cache:
\begin{equation}
    \hat{\mathbf{a}}_{t:t+H}
    \sim
    p_\theta(
        \mathbf{a}_{t:t+H}
        \mid
        o_t, l, \mathcal{C}_{\mathrm{edit}}^{\tau^\star}
    ).
\end{equation}

This inference procedure is more compact than video-generation-based WAMs. 
A video WAM typically denoises and decodes dense spatio-temporal tokens across multiple future frames. 
In contrast, \method{} computes a single set of layer-wise editing caches and uses them directly as context for the action expert. 
Thus, \method{} preserves the reason-before-act principle of WAMs while avoiding the instantiation of dense future-video tokens.

For comparison, we also implement a Fast-WAM-style inference strategy for the video-WAM backbone. 
In this setting, future video tokens are removed at test time. 
The video backbone only processes the current observation and task instruction, and the action expert uses the resulting current-context KV caches for action generation. 
Therefore, this variant keeps a compact action-conditioning interface but avoids future-video token denoising during deployment.

\section{Experiments}
\label{sec:result}

\subsection{Experiment Setup}
\begin{figure}[htbp]
    \centering
    
    \begin{minipage}[t]{0.47\linewidth}
        \centering
        \includegraphics[width=\linewidth]{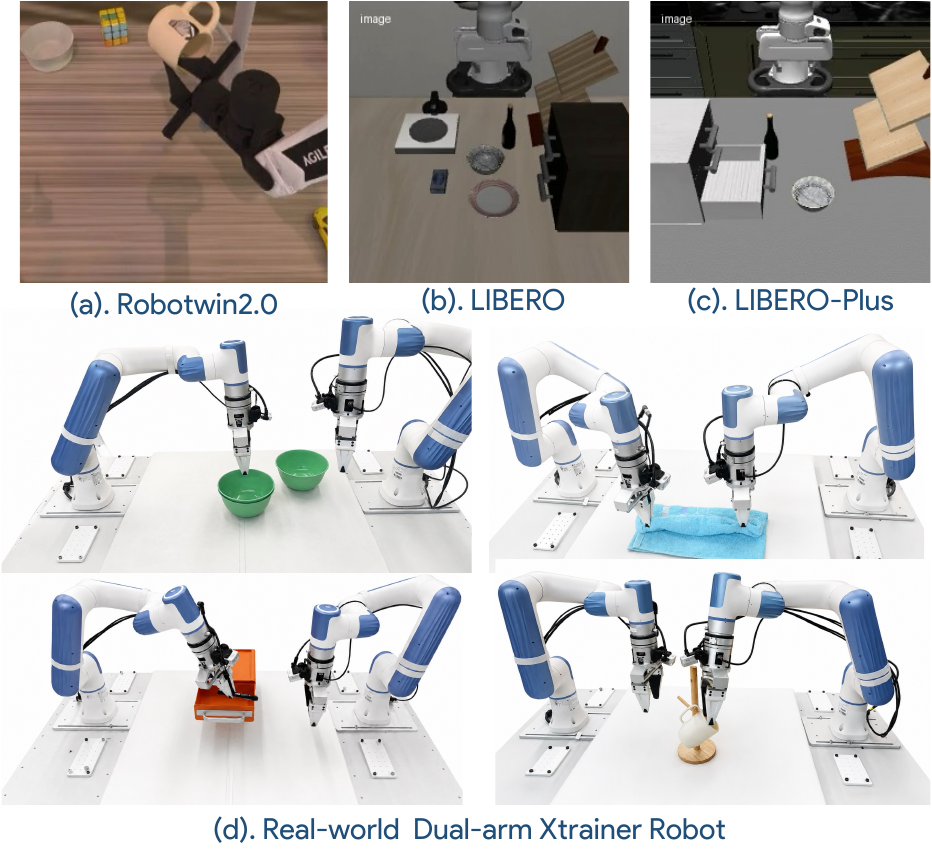}
        \caption{Experiments setup on Robotwin2.0, LIBERO, LIBERO-Plus and real-world robot.}
        \label{fig:exp}
    \end{minipage}
    \hfill
    \begin{minipage}[t]{0.5\linewidth}
        \centering
        \includegraphics[width=\linewidth]{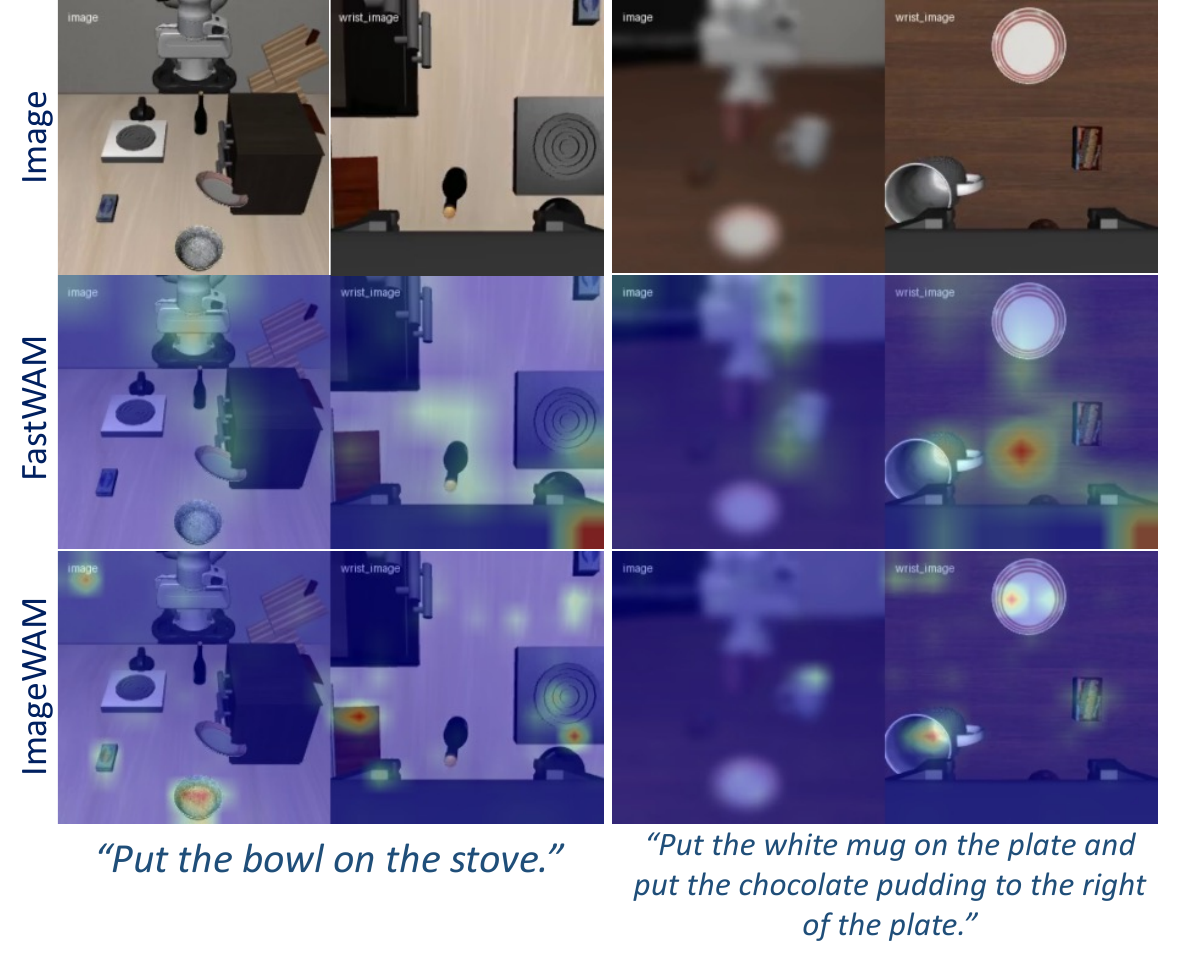}
        \caption{Attention visualization.}
        \label{fig:attention}
    \end{minipage}
    \vspace{-5mm}
\end{figure}

Unlike many VLA and WAM baselines that rely on additional embodied policy pretraining (\textbf{P.T.}), \method{} does not use extra embodied data and is trained only on the downstream benchmark demonstrations.
We evaluate \method{} on LIBERO~\cite{liu2023libero}, LIBERO-Plus~\cite{fei25libero-plus} and RoboTwin 2.0~\cite{chen2025robotwin}, as well as on several real-world manipulation tasks as shown in Figure~\ref{fig:exp} with Flux.2 4B.

\begin{minipage}[htbp]{0.49\textwidth}
\centering
\small
\captionof{table}{Results on RoboTwin2.0.}
\label{tab:robotwin}
\resizebox{\linewidth}{!}{
\begin{tabular}{l|c|cc|c}
\toprule
Method & P.T. & Clean & Rand. & \textbf{Avg.} \\
\midrule
$\pi_{0}$~\cite{24pi0} & \cmark & 65.92 & 58.40 & 62.16 \\
$\pi_{0.5}$~\cite{25pi0.5} & \cmark & 82.74 & 76.76 & 79.75 \\
ABot-M0~\cite{yang2026abot} & \xmark & 81.20 & 80.40 & 80.80 \\
Motus~\cite{bi2025motus} & \cmark & 88.66 & 87.02 & 87.80 \\
LingBot-VA~\cite{li2026lingbotva} & \cmark & 92.90 & 91.50 & 92.20 \\
FastWAM~\cite{yuan2026fastwam} & \xmark & 91.88 & 91.78 & 91.83 \\
\rowcolor{linecolor2}{\textbf{\method{}}} & \xmark & \textbf{93.20} & \textbf{93.56} & \textbf{93.38} \\
\bottomrule
\end{tabular}
}
\end{minipage}
\hfill
\begin{minipage}[t!]{0.50\textwidth}
\centering
\small
\setlength{\tabcolsep}{3pt}
\captionof{table}{Results on LIBERO.}
\label{tab:libero}
\resizebox{\linewidth}{!}{
\begin{tabular}{l|c|cccc|c}
\toprule
Method & P.T. & Spatial & Object & Goal & Long & \textbf{Avg.} \\
\midrule
OpenVLA~\cite{kim2024openvla} & \cmark & 84.7 & 88.4 & 79.2 & 53.7 & 76.5 \\
GR00T N1~\cite{bjorck2025gr00t} & \cmark & 84.7 & 88.4 & 79.2 & 53.7 & 76.5 \\
$\pi_0$~\cite{24pi0} & \cmark & 96.8 & 98.8 & 95.8 & 85.2 & 94.1 \\
$\pi_{0.5}$~\cite{25pi0.5} & \cmark & \textbf{98.8} & 98.2 & \underline{98.0} & 92.4 & 96.9 \\
LingBot-VA~\cite{li2026lingbotva} & \cmark & \underline{98.5} & 99.6 & 97.2 & \underline{98.5} & \textbf{98.5} \\
Motus~\cite{bi2025motus} & \cmark & 96.8 & \underline{99.8} & 96.6 & 97.6 & 97.7 \\
Fast-WAM~\cite{yuan2026fastwam} & \xmark & 98.2 & \textbf{100.0} & 97.0 & 95.2 & 97.6 \\
\rowcolor{linecolor2}{\textbf{\method{}}} & \xmark & 97.2 & 99.2 & \textbf{98.8} & \underline{98.4} & \underline{98.4} \\
\bottomrule
\end{tabular}
}
\end{minipage}

\textbf{LIBERO \& LIBERO-Plus.}
We evaluate our model on LIBERO~\cite{LIBERO23} and LIBERO-Plus~\cite{fei25libero-plus}. 
For LIBERO, we follow the standard benchmarking protocol and train on the four standard suites: Spatial, Object, Goal and LIBERO-Long. 
Each suite contains 500 expert demonstrations spanning 10 tasks. 

LIBERO-Plus provides a more challenging evaluation setting built upon the LIBERO tasks, with increased visual and layout variations. 
Following prior work, we use the same original LIBERO training demonstrations and do not incorporate the augmented LIBERO-Plus training data. 
We evaluate the trained policies under the LIBERO-Plus protocol and report the average success rate.

\textbf{RoboTwin 2.0.}
We further evaluate on RoboTwin 2.0~\cite{chen2025robotwin}, a large-scale simulated benchmark for bimanual robot manipulation. 
The benchmark covers more than 50 tasks and requires policies to coordinate two robot arms under diverse object layouts and scene conditions. 
Following the multi-task setting used in prior work~\cite{li2026lingbotva,yuan2026fastwam}, we train a single policy on demonstrations from all tasks, including 2,500 trajectories collected in clean scenes and 25,000 trajectories collected with heavy scene randomization. 
All models are trained for 30k steps. 
We evaluate each method under both clean and randomized test settings, and report the average success rate over 100 trials per task.

\textbf{Real-world Experiments.}
We also evaluated our model in a real-world dual-arm robot setup. We used the Dobot XTrainer dual-arm robotic platform to collect a dataset consisting of four tasks: \textbf{Stack Three Bowls(T1), Fold Towel(T2), Open Drawer \& Store Marker(T3), and Hang Cup On Rack(T4)}. These tasks involve long-horizon manipulation, visual occlusion, fine-grained manipulation, and deformable-object manipulation, allowing us to assess the real-world performance of the model. Each task contains 100 trajectories. The model was trained on the combined dataset across all tasks, and all models were trained for 30k steps. We report the overall success rate over 100 trials conducted under multiple different initial configurations on this platform.

\subsection{Main Results}

\textbf{Results on RoboTwin 2.0.}
Table~\ref{tab:robotwin} reports the results on RoboTwin 2.0 under both clean and randomized evaluation settings. 
In the clean setting, \method{} achieves an average success rate of 93.20\%. 
In the randomized setting, \method{} achieves an average success rate of 93.56\%. 
Compared with VLA baselines, \method{} shows a clear improvement, indicating that the editing-based world-action context provides useful visual transformation information for multi-task control. 
Compared with video-generation-based WAMs, \method{} reaches comparable performance while avoiding dense future-video token prediction, leading to a more efficient world-action reasoning pathway.

\textbf{Results on LIBERO \& LIBERO-Plus.}
Table~\ref{tab:libero} summarizes the results on LIBERO. 
On the standard LIBERO benchmark, \method{} achieves strong performance across Spatial, Object, Goal, and Long suites, showing that the editing-based backbone is effective for diverse manipulation skills. 
\method{} obtains an average success rate of 98.4\%, remaining competitive with video-generation-based WAMs and pretrained VLA without any data pretraining.

Under the LIBERO-Plus setting, \method{} maintains an average success rate of 83.1\%. 
This suggests that the source-conditioned editing context helps the policy focus on task-relevant visual changes rather than overfitting to fixed visual configurations. 
Together, the results on LIBERO and LIBERO-Plus indicate that image-editing-based world-action reasoning generalizes well across both standard and distribution-shifted simulation benchmarks.

\textbf{Results on Real-world.}
As shown in Table~\ref{tab:realworld}, \method{} achieves an average success rate of 84.5\%, outperforming $\pi_0$ (55.8\%), $\pi_{0.5}$ (72.3\%), and FastWAM (79.0\%). 
Notably, \method{} performs best on all four real-world tasks, covering long-horizon manipulation, deformable-object manipulation, visual occlusion, and fine-grained control. 
Compared with FastWAM, \method{} improves success rates by 6 points on T1 (Stack Three Bowls), 9 points on T2 (Fold Towel), 1 point on T3 (Open Drawer \& Store Marker), and 6 points on T4 (Hang Cup On Rack). 
The largest gain appears on T2, suggesting that the editing-based context is particularly useful when the task requires reasoning about task-relevant visual changes in deformable-object manipulation. 
On T3, both WAM-style methods substantially outperform $\pi_0$, indicating that world-action reasoning helps mitigate the impact of visual occlusion during manipulation. 
Overall, these results show that replacing dense video-token reasoning with image-editing caches yields a practical and efficient WAM backbone.

\begin{table*}[t]
  \caption{Comparison on the LIBERO-Plus benchmark. We report the average success rate across each perturbation dimension, where each perturbation includes the four task suites.}
  \vspace{-2pt}
  \label{table:libero_plus_comparison}
  \begin{center}
    \begin{small}
    \setlength{\tabcolsep}{4pt}
        \begin{tabular}{lccccccccc}
        \toprule
        \multirow{2}{*}{Method} & \multicolumn{9}{c}{LIBERO-Plus}\\
        \cmidrule(lr){2-10}
        & P.T. & Camera & Robot & Language & Light & Background & Noise & Layout & Avg\\
        \midrule
        UniVLA~\cite{bu2025univla} & \cmark & 1.8 & 46.2 & 69.6 & 69.0 & 81.0 & 21.2 & 31.9 & 42.9 \\
        OpenVLA-OFT~\cite{kim2025fine} & \cmark & 56.4 & 31.9 & \cellcolor{linecolor1}{\underline{79.5}} & \cellcolor{linecolor1}{\underline{88.7}} & \cellcolor{linecolor2}{\textbf{93.3}} & 75.8 & 74.2 & 69.6 \\
        $\pi_0$~\cite{24pi0} & \cmark & 13.8 & 6.0 & 58.8 & 85.0 & 81.4 & \cellcolor{linecolor1}{\underline{79.0}} & 68.9 & 53.6 \\
        $\pi_0 \texttt{-Fast} $~\cite{pertsch2025fast} & \cmark & 65.1 & 21.6 & 61.0 & 73.2 & 73.2 & 74.4 & 68.8 & 61.6 \\
        WorldVLA~\cite{cen2025worldvla} & \cmark & 0.1 & 27.9 & 41.6 & 43.7 & 17.1 & 10.9 & 38.0 & 25.0 \\
        FastWAM~\cite{yuan2026fastwam} & \xmark & 16.4 & 44.5 & 68.9 & 78.2 & 53.7 & 37.7 & 60.7 & 51.5 \\
        \midrule
        \textbf{\method{}}(Omnigen2) & \xmark & \cellcolor{linecolor2}{\underline{80.0}} & \cellcolor{linecolor1}{\underline{49.2}} & 70.9 & 82.6 & 69.4 & 77.1 & 71.8 & \cellcolor{linecolor1}{\underline{71.8}} \\
        \textbf{\method{}}(Ovis-U1)  & \xmark & 63.3 & \cellcolor{linecolor2}{\textbf{58.4}} & 75.4 & 86.3 & 66.7 & 75.2 & \cellcolor{linecolor1}{\underline{74.6}} & 71.2 \\
        \textbf{\method{}}(FLUX.2 4B) & \xmark &  \cellcolor{linecolor2}\textbf{80.8} & 50.3 & \cellcolor{linecolor2}\textbf{91.4} & \cellcolor{linecolor2}\textbf{98.1} & \cellcolor{linecolor1}\underline{85.5} & \cellcolor{linecolor2}\textbf{93.8} & \cellcolor{linecolor2}\textbf{80.5} & \cellcolor{linecolor2}\textbf{83.1} \\
        \bottomrule
        \end{tabular}
        \vspace{-6mm}
    \end{small}
  \end{center}
\end{table*}

\begin{table}[t]
  \centering
  \scriptsize
  \begin{minipage}[ht]{0.49\linewidth}
    \centering
    \captionof{table}{
    \textbf{Real-robot eval.}
    Success rates (\%).
    }
    \vspace{-1.5mm}
    \label{tab:realworld}
  \begin{small}
  \setlength{\tabcolsep}{5pt}
    \begin{tabular}{lccccc}
      \toprule
      \textbf{Method} & \textbf{T1} & \textbf{T2} & \textbf{T3} & \textbf{T4} & \textbf{Avg} \\
      \midrule
      $\pi_0$~\cite{24pi0} & 57 & 58 & 54 & 54 & 55.8 \\
      $\pi_{0.5}$~\cite{25pi0.5} & 83 & \underline{77} & 74 & 55 & 72.3 \\
      FastWAM~\cite{yuan2026fastwam} & \underline{88} & 75 & \underline{77} & \underline{76} & \underline{79.0} \\
      \textbf{\method{}(Ours)} & \textbf{94} & \textbf{84} & \textbf{78} & \textbf{82} & \textbf{84.5} \\
      \bottomrule
    \end{tabular}
    \vspace{-2mm}
\end{small}
  \end{minipage}
  \hfill
  \begin{minipage}[ht]{0.49\linewidth}
    \centering
    \captionof{table}{
    \textbf{Efficiency.}
    Lower is better.
    }
    \vspace{-1.5mm}
    \label{tab:efficiency}
    \begin{small}
    \setlength{\tabcolsep}{3pt}
    \begin{tabular}{lccc}
      \toprule
      \textbf{Method} & \textbf{Lat.} & \textbf{TFLOPs} & \textbf{Interm.} \\
      \midrule
      FastWAM-IDM & 1081 ms & 63.65 & Video \\
          FastWAM (1 Step) & 302 ms & 13.21 & Cache \\
      \textbf{\method{}(Ours)} & \textbf{263 ms} & \textbf{9.72} & Cache \\
      \bottomrule
    \end{tabular}
    \end{small}
  \end{minipage}
  \vspace{-2mm}
\end{table}

\subsection{Analysis}

\textbf{Attention Visualization.}
Figure~\ref{fig:attention} visualizes the attention maps from the ImageWAM and FastWAM. 
\method{} concentrates attention on task-relevant change regions, including manipulated objects, target receptacles, and contact areas, while suppressing irrelevant background regions. 
This indicates that the editing caches encode source-grounded and change-centric visual information, providing useful context for the action expert.

\textbf{Latency and FLOPs.}
Table~\ref{tab:efficiency} compares inference latency and FLOPs on A6000 GPU. 
Video-generation WAMs process dense spatio-temporal tokens across multiple future frames, whereas \method{} obtains a single set of image-editing caches from one editing-branch forward step. 
As a result, \method{} reduces latency from 1081 ms to 263 ms and FLOPs from 63.65 to 9.7, while maintaining competitive task success. 
This demonstrates that editing caches offer a more efficient world-action intermediate than future-video token rollout.
\vspace{-2mm}

\begin{figure}[h]
    \centering
    \includegraphics[width=1\linewidth]{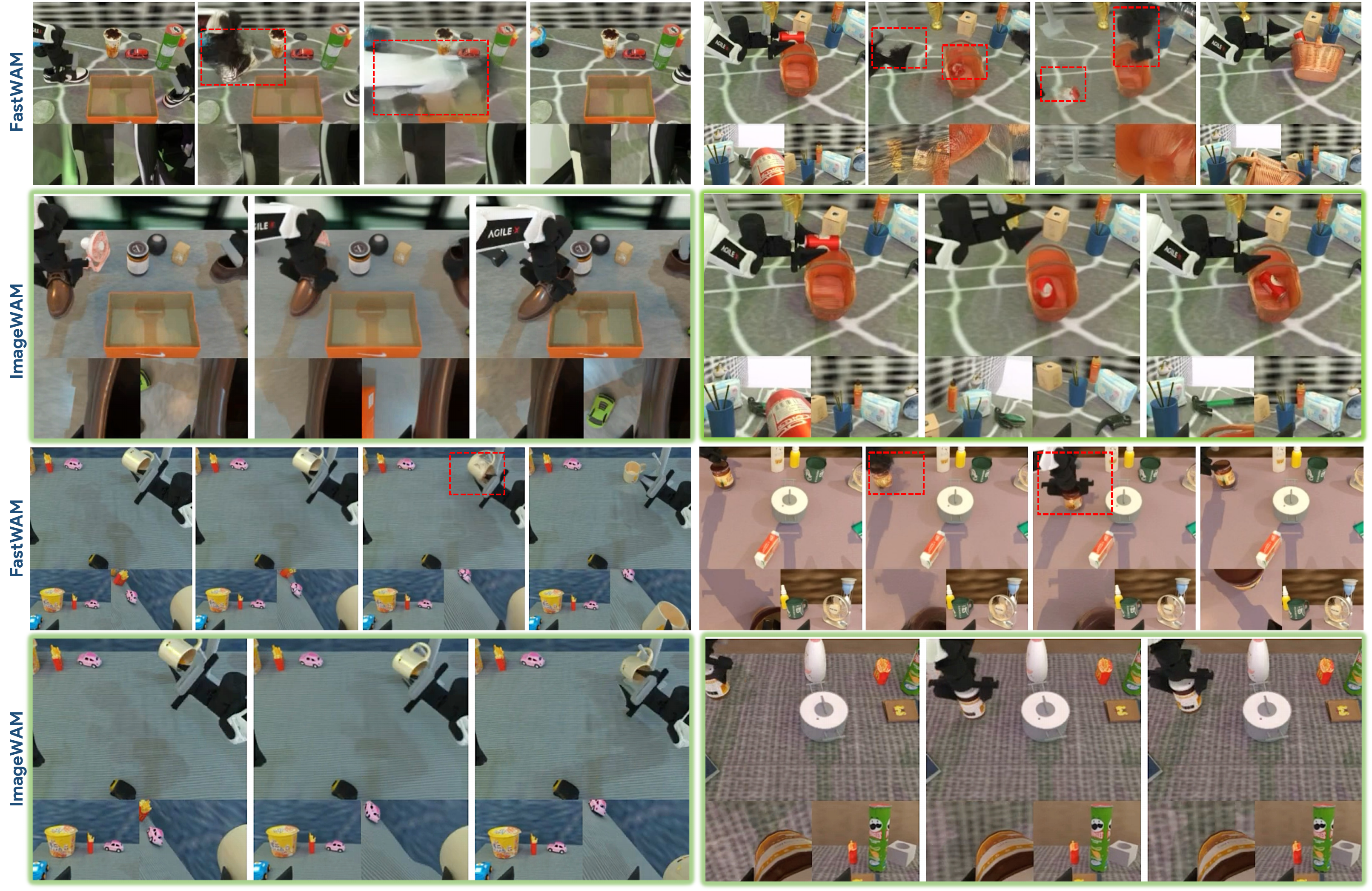}
    \vspace{-8mm}
    \caption{
Future-video artifacts can mislead action prediction. 
The video-WAM baseline generates distorted future observations around task-relevant objects, leading to an unreliable action-conditioning context and task failure. 
\method{} avoids dense imagination and instead conditions the action expert on compact image-editing caches.
}
\vspace{-6mm}
    \label{fig:image_vs_video}
\end{figure}

\paragraph{Qualitative analysis of future-video artifacts.}
Figure~\ref{fig:image_vs_video} illustrates a failure case of video-generation-based WAMs. 
The imagined future frames contain visible artifacts around task-relevant objects, including distorted geometry and inconsistent spatial layout. 
Such artifacts may mislead the action expert, since the predicted action is conditioned on the generated future representation. 
In contrast, \method{} does not instantiate dense future-video tokens or decode future frames at inference time. 
It directly uses image-editing caches as compact action-conditioning context, avoiding the accumulation of visual artifacts in future-video imagination.
\vspace{-2mm}

\vspace{-2mm}
\subsection{Ablation Study}

\question {\bf Can we use different editing models?}
We evaluate whether \method{} depends on a specific editing backbone by replacing OmniGen2~\cite{wu2025omnigen2} with Ovis-U1~\cite{wang2025ovisu1} and FLUX.2 4B~\cite{flux-2-2025}, while keeping the action expert and training data unchanged. 
As shown in Table~\ref{table:libero_plus_comparison}, all variants outperform FastWAM and most VLA baselines on LIBERO-Plus without policy pretraining. 
OmniGen2 and Ovis-U1 achieve similar average success rates of 71.8\% and 71.2\%, respectively, while FLUX.2 4B further improves the average to 83.1\% and performs best on most perturbation dimensions. 
These results show that \method{} is not tied to a particular edit model, and that stronger editing backbones can directly improve policy robustness.

\question{\bf Why do we not use unified understanding-and-generation models?} 

\begin{wraptable}{r}{0.65\linewidth}    
    \vspace{-6mm}
    \captionof{table}{
    \textbf{Comparison with unified understanding-and-generation models.} K.F. denotes keyframe prediction instead of plain future prediction which we adopt.
    }
    \label{tab:umm}
    \vspace{-1.5mm}
  \begin{small}
  \setlength{\tabcolsep}{1.4pt}
    \begin{tabular}{l|c|ccc}
      \toprule
      \textbf{Method} & \textbf{P.T.} & \textbf{LIBERO} & \makecell{\textbf{RoboTwin2.0}\\\textbf{Clean Only}} &  \makecell{\textbf{RoboTwin2.0}\\\textbf{Clean2Hard}}\\
      \midrule
      UniVLA~\cite{wang2025unified} & \cmark & 95.5 & -- & --\\
      BagelVLA (w/ K.F.)~\cite{hu2026bagelvla} & \cmark & -- & 75.3 & \textbf{20.9}\\
      BagelVLA (w/o K.F.)~\cite{hu2026bagelvla} & \cmark & -- & 56.7 & 15.9\\
      \textbf{\method{} (Ours)} & \xmark & \textbf{98.4} & \textbf{84.4} & \underline{18.3}\\
      \bottomrule
    \end{tabular}
  \end{small}
  \vspace{-4mm}
\end{wraptable}

Unified multimodal models that combine understanding and generation are promising, but the two capabilities impose different architectural demands. 
Understanding benefits from high-level semantic abstraction, whereas generation requires fine-grained spatial and structural details, especially in deeper layers~\cite{yu2025repa}. 
Jointly optimizing both objectives in a single fully shared model may therefore introduce interference, where improving generation can hurt understanding, and vice versa. 
Instead, \method{} decouples these roles: we keep the VLM-based understanding components frozen and adapt only the diffusion generation branch and the action expert for robot control. 
As shown in Table~\ref{tab:umm}, this design outperforms UniVLA and BagelVLA under similar non-keyframe future prediction setting, which are built upon unified understanding-and-generation models, while requiring no additional policy pretraining.

\question{\bf What is the effect of the size of the editing backbone?} We evaluate whether increasing the capacity of the editing backbone improves the robustness of the policy in LIBERO-Plus. Replacing FLUX.2 4B with a larger FLUX.2 backbone increases the average success rate from 83.1\% to 85.21\%. The improvement mainly comes from Robot, Language, Background, and Layout perturbations, suggesting that larger editing models provide stronger instruction-conditioned visual context for action prediction. However, the gains are not uniform across all dimensions: Camera, Light, and Noise do not improve monotonically. This indicates that backbone scaling generally improves robustness, but the benefit depends on how the editing cache aligns with different perturbation types.

\begin{table*}[h]
\vspace{-2mm}
\caption{Effect of using a larger editing backbone on LIBERO-Plus. We report the average success rate across each perturbation dimension, where each dimension includes the four LIBERO task suites.}
  \vspace{-4pt}
  \label{table:libero_plus_comparison}
  \begin{center}
    \begin{small}
    \setlength{\tabcolsep}{3pt}
        \begin{tabular}{lccccccccc}
        \toprule
        \multirow{2}{*}{Method} & \multicolumn{9}{c}{LIBERO-Plus}\\
        \cmidrule(lr){2-10}
        & P.T. & Camera & Robot & Language & Light & Background & Noise & Layout & Avg\\
        \midrule
        \method{(FLUX.2 4B)} & \xmark &  80.8 & 50.3 & 91.4 & 98.1 & 85.5 & 93.8 & 80.5 & 83.1 \\
        \method{(FLUX.2 9B)} & \xmark & 79.8 & 58.7 & 95.2 & 96.1 & 91.2 & 93.3 & 83.1 & 85.2 \\
        \bottomrule
        \end{tabular}
        \vspace{-6mm}
    \end{small}
  \end{center}
\end{table*}

\section{Conclusion}
\label{sec:conclusion}
\vspace{-1mm}

    In this paper, we explore employing an image editing rather than a video generation model as the WAM backbone because image editing is an inherently ideal general task that naturally demands both visual understanding and generation.
    By simply predicting a single future frame, our model provides strong intermediate representations for the action model and enables end-to-end policy learning. 
    Our model achieves a 93.56\% success rate on RoboTwin (Random), substantially outperforming all VLA baselines and reaching performance comparable to state-of-the-art WAM models. 
    We argue that the language-vision interaction priors in editing models drive our model’s effectiveness and lay the groundwork for broader use of image models.

\clearpage

\setlength{\bibsep}{5pt}
\bibliography{main}

@String(ICLR = {Int. Conf. Learn. Represent.})

@String(AAAI = {AAAI})

@String(ICLR  = {ICLR})

@String(NeurIPS  = {NeurIPS})

@article{du2024learning,
  title={Learning universal policies via text-guided video generation},
  author={Du, Yilun and Yang, Sherry and Dai, Bo and Dai, Hanjun and Nachum, Ofir and Tenenbaum, Josh and Schuurmans, Dale and Abbeel, Pieter},
  journal={NeurIPS},
  year={2024}
}

@article{kim2024openvla,
  title={OpenVLA: An Open-Source Vision-Language-Action Model},
  author={Kim, Moo Jin and Pertsch, Karl and Karamcheti, Siddharth and Xiao, Ted and Balakrishna, Ashwin and Nair, Suraj and Rafailov, Rafael and Foster, Ethan and Lam, Grace and Sanketi, Pannag and others},
  journal={arXiv preprint},
  year={2024}
}

@article{liu2023libero,
  title={LIBERO: Benchmarking Knowledge Transfer for Lifelong Robot Learning},
  author={Liu, Bo and Zhu, Yifeng and Gao, Chongkai and Feng, Yihao and Liu, Qiang and Zhu, Yuke and Stone, Peter},
  journal={arXiv preprint},
  year={2023}
}

@article{susie23,
  title={Zero-shot robotic manipulation with pretrained image-editing diffusion models},
  author={Black, Kevin and Nakamoto, Mitsuhiko and Atreya, Pranav and Walke, Homer and Finn, Chelsea and Kumar, Aviral and Levine, Sergey},
  journal={arXiv preprint},
  year={2023}
}

@article{seer24,
  title={Predictive inverse dynamics models are scalable learners for robotic manipulation},
  author={Tian, Yang and Yang, Sizhe and Zeng, Jia and Wang, Ping and Lin, Dahua and Dong, Hao and Pang, Jiangmiao},
  journal=ICLR,
  year={2024}
}

@article{cotvla25,
  title={Cot-vla: Visual chain-of-thought reasoning for vision-language-action models},
  author={Zhao, Qingqing and Lu, Yao and Kim, Moo Jin and Fu, Zipeng and Zhang, Zhuoyang and Wu, Yecheng and Li, Zhaoshuo and Ma, Qianli and Han, Song and Finn, Chelsea and others},
  journal={arXiv preprint},
  year={2025}
}

@article{vpp2024hu,
  title={Video Prediction Policy: A Generalist Robot Policy with Predictive Visual Representations},
  author={Hu, Yucheng and Guo, Yanjiang and Wang, Pengchao and Chen, Xiaoyu and Wang, Yen-Jen and Zhang, Jianke and Sreenath, Koushil and Lu, Chaochao and Chen, Jianyu},
  journal={arXiv preprint},
  year={2024}
}

@inproceedings{LIBERO23,
  author       = {Bo Liu and
                  Yifeng Zhu and
                  Chongkai Gao and
                  Yihao Feng and
                  Qiang Liu and
                  Yuke Zhu and
                  Peter Stone},
  editor       = {Alice Oh and
                  Tristan Naumann and
                  Amir Globerson and
                  Kate Saenko and
                  Moritz Hardt and
                  Sergey Levine},
  title        = {{LIBERO:} Benchmarking Knowledge Transfer for Lifelong Robot Learning},
  booktitle    = {NeurIPS},
  year         = {2023},
}

@article{24pi0,
  title={pi0: A Vision-Language-Action Flow Model for General Robot Control},
  author={Black, Kevin and Brown, Noah and Driess, Danny and Esmail, Adnan and Equi, Michael and Finn, Chelsea and Fusai, Niccolo and Groom, Lachy and Hausman, Karol and Ichter, Brian and others},
  journal={arXiv preprint},
  year={2024}
}

@article{25pi0.5,
  title={pi0.5: a Vision-Language-Action Model with Open-World Generalization},
  author={Intelligence, Physical and Black, Kevin and Brown, Noah and Darpinian, James and Dhabalia, Karan and Driess, Danny and Esmail, Adnan and Equi, Michael and Finn, Chelsea and Fusai, Niccolo and others},
  journal={arXiv preprint},
  year={2025}
}

@article{qu25spatialvla,
  title={Spatialvla: Exploring spatial representations for visual-language-action model},
  author={Qu, Delin and Song, Haoming and Chen, Qizhi and Yao, Yuanqi and Ye, Xinyi and Ding, Yan and Wang, Zhigang and Gu, JiaYuan and Zhao, Bin and Wang, Dong and others},
  journal={arXiv preprint},
  year={2025}
}

@article{kim2025fine,
  title={Fine-tuning vision-language-action models: Optimizing speed and success},
  author={Kim, Moo Jin and Finn, Chelsea and Liang, Percy},
  journal={arXiv preprint},
  year={2025}
}

@article{bu2025univla,
  title={UniVLA: Learning to Act Anywhere with Task-centric Latent Actions},
  author={Bu, Qingwen and Yang, Yanting and Cai, Jisong and Gao, Shenyuan and Ren, Guanghui and Yao, Maoqing and Luo, Ping and Li, Hongyang},
  journal={arXiv preprint},
  year={2025}
}

@article{zhu2025unified,
  title={Unified World Models: Coupling Video and Action Diffusion for Pretraining on Large Robotic Datasets},
  author={Zhu, Chuning and Yu, Raymond and Feng, Siyuan and Burchfiel, Benjamin and Shah, Paarth and Gupta, Abhishek},
  journal={arXiv preprint},
  year={2025}
}

@article{bjorck2025gr00t,
  title={Gr00t n1: An open foundation model for generalist humanoid robots},
  author={Bjorck, Johan and Casta{\~n}eda, Fernando and Cherniadev, Nikita and Da, Xingye and Ding, Runyu and Fan, Linxi and Fang, Yu and Fox, Dieter and Hu, Fengyuan and Huang, Spencer and others},
  journal={arXiv preprint},
  year={2025}
}

@article{pertsch2025fast,
  title={Fast: Efficient action tokenization for vision-language-action models},
  author={Pertsch, Karl and Stachowicz, Kyle and Ichter, Brian and Driess, Danny and Nair, Suraj and Vuong, Quan and Mees, Oier and Finn, Chelsea and Levine, Sergey},
  journal={arXiv preprint},
  year={2025}
}

@article{zhang2025dreamvla,
  title={DreamVLA: A Vision-Language-Action Model Dreamed with Comprehensive World Knowledge},
  author={Zhang, Wenyao and Liu, Hongsi and Qi, Zekun and Wang, Yunnan and Yu, Xinqiang and Zhang, Jiazhao and Dong, Runpei and He, Jiawei and Wang, He and Zhang, Zhizheng and others},
  journal={arXiv preprint},
  year={2025}
}

@article{feng2025vidar,
  title={Generalist Bimanual Manipulation via Foundation Video Diffusion Models},
  author={Feng, Yao and Tan, Hengkai and Mao, Xinyi and Liu, Guodong and Huang, Shuhe and Xiang, Chendong and Su, Hang and Zhu, Jun},
  journal={arXiv preprint},
  year={2025}
}

@article{wen2024vidman,
  title={Vidman: Exploiting implicit dynamics from video diffusion model for effective robot manipulation},
  author={Wen, Youpeng and Lin, Junfan and Zhu, Yi and Han, Jianhua and Xu, Hang and Zhao, Shen and Liang, Xiaodan},
  journal={NeurIPS},
  year={2024}
}

@article{cen2025worldvla,
  title={WorldVLA: Towards Autoregressive Action World Model},
  author={Cen, Jun and Yu, Chaohui and Yuan, Hangjie and Jiang, Yuming and Huang, Siteng and Guo, Jiayan and Li, Xin and Song, Yibing and Luo, Hao and Wang, Fan and others},
  journal={arXiv preprint},
  year={2025}
}

@article{tan2025anypos,
  title={AnyPos: Automated Task-Agnostic Actions for Bimanual Manipulation},
  author={Tan, Hengkai and Feng, Yao and Mao, Xinyi and Huang, Shuhe and Liu, Guodong and Hao, Zhongkai and Su, Hang and Zhu, Jun},
  journal={arXiv preprint},
  year={2025}
}

@article{wu2026pragmatic,
  title={A Pragmatic VLA Foundation Model},
  author={Wu, Wei and Lu, Fan and Wang, Yunnan and Yang, Shuai and Liu, Shi and Wang, Fangjing and Zhu, Qian and Sun, He and Wang, Yong and Ma, Shuailei and others},
  journal={arXiv preprint arXiv:2601.18692},
  year={2026}
}

@article{sun2026vla,
  title={VLA-JEPA: Enhancing Vision-Language-Action Model with Latent World Model},
  author={Sun, Jingwen and Zhang, Wenyao and Qi, Zekun and Ren, Shaojie and Liu, Zezhi and Zhu, Hanxin and Sun, Guangzhong and Jin, Xin and Chen, Zhibo},
  journal={arXiv preprint arXiv:2602.10098},
  year={2026}
}

@article{zhang2026disentangled,
  title={Disentangled Robot Learning via Separate Forward and Inverse Dynamics Pretraining},
  author={Zhang, Wenyao and Zhang, Bozhou and Qi, Zekun and Zeng, Wenjun and Jin, Xin and Zhang, Li},
  journal={arXiv preprint arXiv:2604.16391},
  year={2026}
}

@article{li2026world,
  title={World-Value-Action Model: Implicit Planning for Vision-Language-Action Systems},
  author={Li, Runze and Zhang, Hongyin and Jin, Junxi and Zeng, Qixin and Zhuang, Zifeng and Tang, Yiqi and Lyu, Shangke and Wang, Donglin},
  journal={arXiv preprint arXiv:2604.14732},
  year={2026}
}

@article{ye2026dreamzero,
  title={World action models are zero-shot policies},
  author={Ye, Seonghyeon and Ge, Yunhao and Zheng, Kaiyuan and Gao, Shenyuan and Yu, Sihyun and Kurian, George and Indupuru, Suneel and Tan, You Liang and Zhu, Chuning and Xiang, Jiannan and others},
  journal={arXiv preprint arXiv:2602.15922},
  year={2026}
}

@article{li2026lingbotva,
  title={Causal World Modeling for Robot Control},
  author={Li, Lin and Zhang, Qihang and Luo, Yiming and Yang, Shuai and Wang, Ruilin and Han, Fei and Yu, Mingrui and Gao, Zelin and Xue, Nan and Zhu, Xing and others},
  journal={arXiv preprint arXiv:2601.21998},
  year={2026}
}

@article{yuan2026fastwam,
  title={Fast-WAM: Do World Action Models Need Test-time Future Imagination?},
  author={Yuan, Tianyuan and Dong, Zibin and Liu, Yicheng and Zhao, Hang},
  journal={arXiv preprint arXiv:2603.16666},
  year={2026}
}

@article{ye2026gigaworldpolicy,
  title={GigaWorld-Policy: An Efficient Action-Centered World--Action Model},
  author={Ye, Angen and Wang, Boyuan and Ni, Chaojun and Huang, Guan and Zhao, Guosheng and Li, Hao and Li, Hengtao and Li, Jie and Lv, Jindi and Liu, Jingyu and others},
  journal={arXiv preprint arXiv:2603.17240},
  year={2026}
}

@article{hu2026bagelvla,
  title={BagelVLA: Enhancing Long-Horizon Manipulation via Interleaved Vision-Language-Action Generation},
  author={Hu, Yucheng and Zhang, Jianke and Luo, Yuanfei and Guo, Yanjiang and Chen, Xiaoyu and Sun, Xinshu and Feng, Kun and Lu, Qingzhou and Chen, Sheng and Zhang, Yangang and others},
  journal={arXiv preprint arXiv:2602.09849},
  year={2026}
}

@article{bi2025motus,
  title={Motus: A unified latent action world model},
  author={Bi, Hongzhe and Tan, Hengkai and Xie, Shenghao and Wang, Zeyuan and Huang, Shuhe and Liu, Haitian and Zhao, Ruowen and Feng, Yao and Xiang, Chendong and Rong, Yinze and others},
  journal={arXiv preprint arXiv:2512.13030},
  year={2025}
}

@article{madit4dit,
  title={DIT4DIT: JOINTLY MODELING VIDEO DYNAMICS AND ACTIONS FOR GENERALIZABLE ROBOT CONTROL},
  author={Ma, Teli and Zheng, Jia and Wang, Zifan and Jiang, Chunli and Cui, Andy and Liang, Junwei and Yang, Shuo},
  journal={arXiv preprint arXiv:2603.10448},
  year={2025}
}

@article{lyu2026lda,
  title={LDA-1B: Scaling Latent Dynamics Action Model via Universal Embodied Data Ingestion},
  author={Lyu, Jiangran and Liu, Kai and Zhang, Xuheng and Liao, Haoran and Feng, Yusen and Zhu, Wenxuan and Shen, Tingrui and Chen, Jiayi and Zhang, Jiazhao and Dong, Yifei and others},
  journal={arXiv preprint arXiv:2602.12215},
  year={2026}
}

@article{li2025worldeval,
  title={Worldeval: World model as real-world robot policies evaluator},
  author={Li, Yaxuan and Zhu, Yichen and Wen, Junjie and Shen, Chaomin and Xu, Yi},
  journal={arXiv preprint arXiv:2505.19017},
  year={2025}
}

@article{yang2025orv,
  title={Orv: 4d occupancy-centric robot video generation},
  author={Yang, Xiuyu and Li, Bohan and Xu, Shaocong and Wang, Nan and Ye, Chongjie and Chen, Zhaoxi and Qin, Minghan and Ding, Yikang and Zhu, Zheng and Jin, Xin and others},
  journal={arXiv preprint arXiv:2506.03079},
  year={2025}
}

@article{zhen2025tesseract,
  title={TesserAct: Learning 4D Embodied World Models}, 
  author={Haoyu Zhen and Qiao Sun and Hongxin Zhang and Junyan Li and Siyuan Zhou and Yilun Du and Chuang Gan},
  year={2025},
  eprint={2504.20995},
  archivePrefix={arXiv},
  primaryClass={cs.CV},
  url={https://arxiv.org/abs/2504.20995}, 
}

@article{peng2026reworld,
  title={ReWorld: Multi-Dimensional Reward Modeling for Embodied World Models},
  author={Peng, Baorui and Zhang, Wenyao and Xu, Liang and Qi, Zekun and Zhang, Jiazhao and Liu, Hongsi and Zeng, Wenjun and Jin, Xin},
  journal={arXiv preprint arXiv:2601.12428},
  year={2026}
}

@article{xu2026kinema4d,
  title={Kinema4D: Kinematic 4D World Modeling for Spatiotemporal Embodied Simulation},
  author={Xu, Mutian and Zhang, Tianbao and Liu, Tianqi and Chen, Zhaoxi and Han, Xiaoguang and Liu, Ziwei},
  journal={arXiv preprint arXiv:2603.16669},
  year={2026}
}

@article{jiang2026wovr,
  title={Wovr: World models as reliable simulators for post-training vla policies with rl},
  author={Jiang, Zhennan and Zhou, Shangqing and Jiang, Yutong and Huang, Zefang and Wei, Mingjie and Chen, Yuhui and Zhou, Tianxing and Guo, Zhen and Lin, Hao and Zhang, Quanlu and others},
  journal={arXiv preprint arXiv:2602.13977},
  year={2026}
}

@article{zhang2026robostereo,
  title={RoboStereo: Dual-Tower 4D Embodied World Models for Unified Policy Optimization},
  author={Zhang, Ruicheng and Chen, Guangyu and Xu, Zunnan and Liu, Zihao and Zhong, Zhizhou and Zhang, Mingyang and Zhou, Jun and Li, Xiu},
  journal={arXiv preprint arXiv:2603.12639},
  year={2026}
}

@article{chen2026unit,
  title={UniT: Toward a Unified Physical Language for Human-to-Humanoid Policy Learning and World Modeling},
  author={Chen, Boyu and Chen, Yi and Qiu, Lu and Bai, Jerry and Ge, Yuying and Ge, Yixiao},
  journal={arXiv preprint arXiv:2604.19734},
  year={2026}
}

@article{bardhan2026persistent,
  title={Persistent Robot World Models: Stabilizing Multi-Step Rollouts via Reinforcement Learning},
  author={Bardhan, Jai and Drozdik, Patrik and Sivic, Josef and Petrik, Vladimir},
  journal={arXiv preprint arXiv:2603.25685},
  year={2026}
}

@article{wei2026fate,
  title={FATE: Closed-Loop Feasibility-Aware Task Generation with Active Repair for Physically Grounded Robotic Curricula},
  author={Wei, Bingchuan and Huang, Bingqi and Ma, Jingheng and Cui, Sen and others},
  journal={arXiv preprint arXiv:2603.01505},
  year={2026}
}

@article{lang2026vag,
  title={VAG: Dual-Stream Video-Action Generation for Embodied Data Synthesis},
  author={Lang, Xiaolei and Wang, Yang and Zhou, Yukun and Ni, Chaojun and Li, Kerui and Zhu, Jiagang and Liu, Tianze and Lv, Jiajun and Zuo, Xingxing and Ye, Yun and others},
  journal={arXiv preprint arXiv:2604.09330},
  year={2026}
}

@article{wang2026interactive,
  title={Interactive World Simulator for Robot Policy Training and Evaluation},
  author={Wang, Yixuan and Syed, Rhythm and Wu, Fangyu and Zhang, Mengchao and Onol, Aykut and Barreiros, Jose and Nayyeri, Hooshang and Dear, Tony and Zhang, Huan and Li, Yunzhu},
  journal={arXiv preprint arXiv:2603.08546},
  year={2026}
}

@article{liu2026world,
  title={World Action Verifier: Self-Improving World Models via Forward-Inverse Asymmetry},
  author={Liu, Yuejiang and Feng, Fan and Kong, Lingjing and Lu, Weifeng and Tang, Jinzhou and Zhang, Kun and Murphy, Kevin and Finn, Chelsea and Du, Yilun},
  journal={arXiv preprint arXiv:2604.01985},
  year={2026}
}

@article{Lv2025F1AV,
  title={F1: A Vision-Language-Action Model Bridging Understanding and Generation to Actions},
  author={Qi Lv and Weijie Kong and Hao Li and Jia Zeng and Zherui Qiu and Delin Qu and Haoming Song and Qizhi Chen and Xiang Deng and Jiangmiao Pang},
  journal={ArXiv},
  year={2025},
  volume={abs/2509.06951},
  url={https://api.semanticscholar.org/CorpusID:281204333}
}

@inproceedings{Liu2026WorldAV,
  title={World Action Verifier: Self-Improving World Models via Forward-Inverse Asymmetry},
  author={Yuejiang Liu and Fan Feng and Lingjing Kong and Weifeng Lu and Jinzhou Tang and Kun Zhang and Kevin P. Murphy and Chelsea Finn and Yilun Du},
  year={2026},
  url={https://api.semanticscholar.org/CorpusID:287074218}
}

@article{Chen2025LargeVP,
  title={Large Video Planner Enables Generalizable Robot Control},
  author={Boyuan Chen and Tianyuan Zhang and Haoran Geng and Kiwhan Song and Caiyi Zhang and Peihao Li and William T. Freeman and Jitendra Malik and Pieter Abbeel and Russ Tedrake and Vincent Sitzmann and Yilun Du},
  journal={ArXiv},
  year={2025},
  volume={abs/2512.15840},
  url={https://api.semanticscholar.org/CorpusID:283933826}
}

@inproceedings{Li2026dWorldEvalSR,
  title={dWorldEval: Scalable Robotic Policy Evaluation via Discrete Diffusion World Model},
  author={Yaxuan Li and Zhongyi Zhou and Ye Chen and Yaokai Xue and Yichen Zhu},
  year={2026},
  url={https://api.semanticscholar.org/CorpusID:287773839}
}

@article{Liao2025GenieEA,
  title={Genie Envisioner: A Unified World Foundation Platform for Robotic Manipulation},
  author={Yue Liao and Yue Liao and Pengfei Zhou and Siyuan Huang and Donglin Yang and Shengcong Chen and Yuxin Jiang and Hu Yue and Jingbin Cai and Si Liu and Jianlan Luo and Liliang Chen and Shuicheng Yan and Maoqing Yao and Guanghui Ren},
  journal={ArXiv},
  year={2025},
  volume={abs/2508.05635},
  url={https://api.semanticscholar.org/CorpusID:280545868}
}

@inproceedings{Wang2026InteractiveWS,
  title={Interactive World Simulator for Robot Policy Training and Evaluation},
  author={Yixuan Wang and Rhythm Syed and Fangyu Wu and Mengchao Zhang and Aykut Onol and Jose Barreiros and Hooshang Nayyeri and Tony Dear and Huan Zhang and Yunzhu Li},
  year={2026},
  url={https://api.semanticscholar.org/CorpusID:286377674}
}

@article{Mi2026TCIDMGV,
  title={TC-IDM: Grounding Video Generation for Executable Zero-shot Robot Motion},
  author={Weishi Mi and Yong Bao and Xiaowei Chi and Xiaozhu Ju and Zhiyuan Qin and Kuangzhi Ge and Kai Tang and Peidong Jia and Shanghang Zhang and Jian Tang},
  journal={ArXiv},
  year={2026},
  volume={abs/2601.18323},
  url={https://api.semanticscholar.org/CorpusID:285051517}
}

@inproceedings{Zhang2026VeoActHF,
  title={Veo-Act: How Far Can Frontier Video Models Advance Generalizable Robot Manipulation?},
  author={Zhongrui Zhang and Cheng‐Chuan Yang and Qin Lu and Yanjiang Guo and Jianke Zhang and Yucheng Hu and Jianyu Chen},
  year={2026},
  url={https://api.semanticscholar.org/CorpusID:287202336}
}

@article{kim2026cosmos,
  title={Cosmos policy: Fine-tuning video models for visuomotor control and planning},
  author={Kim, Moo Jin and Gao, Yihuai and Lin, Tsung-Yi and Lin, Yen-Chen and Ge, Yunhao and Lam, Grace and Liang, Percy and Song, Shuran and Liu, Ming-Yu and Finn, Chelsea and others},
  journal={arXiv preprint arXiv:2601.16163},
  year={2026}
}

@article{wang2024scene,
  title={Scene graph disentanglement and composition for generalizable complex image generation},
  author={Wang, Yunnan and Li, Ziqiang and Zhang, Wenyao and Zhang, Zequn and Xie, Baao and Liu, Xihui and Zeng, Wenjun and Jin, Xin},
  journal={Advances in Neural Information Processing Systems},
  volume={37},
  pages={98478--98504},
  year={2024}
}

@article{zheng2026uni,
  title={Uni-Edit: Intelligent Editing Is A General Task For Unified Model Tuning},
  author={Zheng, Dian and Zhang, Manyuan and Li, Hongyu and Liu, Hongbo and Zou, Kai and Feng, Kaituo and Li, Hongsheng},
  journal={arXiv preprint arXiv:2605.21487},
  year={2026}
}

@article{team2026hy,
  title={HY-Embodied-0.5: Embodied Foundation Models for Real-World Agents},
  author={Team, HY and Yu, Xumin and Liu, Zuyan and Wang, Ziyi and Zhang, He and Rao, Yongming and Liu, Fangfu and Zhang, Yani and Zhao, Ruowen and Wang, Oran and others},
  journal={arXiv preprint arXiv:2604.07430},
  year={2026}
}

@article{lin2026universal,
  title={Universal Pose Pretraining for Generalizable Vision-Language-Action Policies},
  author={Lin, Haitao and Yu, Hanyang and Huang, Jingshun and Zhang, He and Ling, Yonggen and Tan, Ping and Xue, Xiangyang and Fu, Yanwei},
  journal={arXiv preprint arXiv:2602.19710},
  year={2026}
}

@inproceedings{zhang2023magicbrush,
  title={MagicBrush: A Manually Annotated Dataset for Instruction-Guided Image Editing},
  author={Zhang, Kai and Mo, Lingbo and Chen, Wenhu and Sun, Huan and Su, Yu},
  booktitle={Advances in Neural Information Processing Systems},
  year={2023}
}

@inproceedings{fu2024mgie,
  title={Guiding Instruction-Based Image Editing via Multimodal Large Language Models},
  author={Fu, Tsu-Jui and Hu, Wenze and Du, Xianzhi and Wang, William Yang and Yang, Yinfei and Gan, Zhe},
  booktitle={International Conference on Learning Representations},
  year={2024}
}

@inproceedings{sheynin2024emuedit,
  title={Emu Edit: Precise Image Editing via Recognition and Generation Tasks},
  author={Sheynin, Shelly and Polyak, Adam and Singer, Uriel and Kirstain, Yuval and Zohar, Amit and Ashual, Oron and Parikh, Devi and Taigman, Yaniv},
  booktitle={Proceedings of the IEEE/CVF Conference on Computer Vision and Pattern Recognition},
  pages={8871--8879},
  year={2024}
}

@inproceedings{yu2025anyedit,
  title={AnyEdit: Mastering Unified High-Quality Image Editing for Any Idea},
  author={Yu, Qifan and Chow, Wei and Yue, Zhongqi and Pan, Kaihang and Wu, Yang and Wan, Xiaoyang and Li, Juncheng and Tang, Siliang and Zhang, Hanwang and Zhuang, Yueting},
  booktitle={Proceedings of the IEEE/CVF Conference on Computer Vision and Pattern Recognition},
  pages={26125--26135},
  year={2025}
}

@misc{gpt1p5,
    author={OpenAI},
    title={{GPT-Image-1.5}},
      year         = {2026},
      howpublished = {\url{https://openai.com/index/new-chatgpt-images-is-here/}},
      note         = {Accessed: 2026-03-19}
}

@article{wu2025qwen,
  title={Qwen-image technical report},
  author={Wu, Chenfei and Li, Jiahao and Zhou, Jingren and Lin, Junyang and Gao, Kaiyuan and Yan, Kun and Yin, Sheng-ming and Bai, Shuai and Xu, Xiao and Chen, Yilei and others},
  journal={arXiv preprint arXiv:2508.02324},
  year={2025}
}

@article{chen2025robotwin,
  title={Robotwin 2.0: A scalable data generator and benchmark with strong domain randomization for robust bimanual robotic manipulation},
  author={Chen, Tianxing and Chen, Zanxin and Chen, Baijun and Cai, Zijian and Liu, Yibin and Li, Zixuan and Liang, Qiwei and Lin, Xianliang and Ge, Yiheng and Gu, Zhenyu and others},
  journal={arXiv preprint arXiv:2506.18088},
  year={2025}
}

@article{zhang2026uam,
  title   = {UAM: A Dual-Stream Perspective on Forgetting in VLA Training},
  author  = {Zhang, Jianke and Luo, Yuanfei and Hu, Yucheng and Chen, Xiaoyu and Guo, Yanjiang and Liu, Ziyang and Xu, Hongbin and Lan, Tian and Chen, Jianyu},
  journal = {arXiv preprint arXiv:2605.15735},
  year    = {2026}
}

@article{fei25libero-plus,
    title={LIBERO-Plus: In-depth Robustness Analysis of Vision-Language-Action Models},
    author={Senyu Fei and Siyin Wang and Junhao Shi and Zihao Dai and Jikun Cai and Pengfang Qian and Li Ji and Xinzhe He and Shiduo Zhang and Zhaoye Fei and Jinlan Fu and Jingjing Gong and Xipeng Qiu},
    journal = {arXiv preprint arXiv:2510.13626},
    year={2025},
}

@article{wu2025omnigen2,
  title={OmniGen2: Exploration to Advanced Multimodal Generation},
  author={Chenyuan Wu and Pengfei Zheng and Ruiran Yan and Shitao Xiao and Xin Luo and Yueze Wang and Wanli Li and Xiyan Jiang and Yexin Liu and Junjie Zhou and Ze Liu and Ziyi Xia and Chaofan Li and Haoge Deng and Jiahao Wang and Kun Luo and Bo Zhang and Defu Lian and Xinlong Wang and Zhongyuan Wang and Tiejun Huang and Zheng Liu},
  journal={arXiv preprint arXiv:2506.18871},
  year={2025}
}

@article{wang2025ovisu1,
  title={Ovis-U1 Technical Report}, 
  author={Wang, Guo-Hua and Zhao, Shanshan and Zhang, Xinjie and Cao, Liangfu and Zhan, Pengxin and Duan, Lunhao and Lu, Shiyin and Fu, Minghao and Zhao, Jianshan and Li, Yang and Chen, Qing-Guo},
  journal={arXiv preprint arXiv:2506.23044},
  year={2025}
}

@misc{flux-2-2025,
    author={Black Forest Labs},
    title={{FLUX.2: Frontier Visual Intelligence}},
    year={2025},
    howpublished={\url{https://bfl.ai/blog/flux-2}},
}

@article{yuan2025depthvla,
  title={Depthvla: Enhancing vision-language-action models with depth-aware spatial reasoning},
  author={Yuan, Tianyuan and Liu, Yicheng and Lu, Chenhao and Chen, Zhuoguang and Jiang, Tao and Zhao, Hang},
  journal={arXiv preprint arXiv:2510.13375},
  year={2025}
}

@article{ge2026vampo,
  title={VAMPO: Policy Optimization for Improving Visual Dynamics in Video Action Models},
  author={Ge, Zirui and Ding, Pengxiang and Yin, Baohua and Wang, Qishen and Xie, Zhiyong and Wang, Yemin and Wang, Jinbo and Li, Hengtao and Suo, Runze and Song, Wenxuan and others},
  journal={arXiv preprint arXiv:2603.19370},
  year={2026}
}

@inproceedings{beingbeyond2025beingh0,
  title={Being-H0: Vision-Language-Action Pretraining from Large-Scale Human Videos},
  author={Luo, Hao and Feng, Yicheng and Zhang, Wanpeng and Zheng, Sipeng and Wang, Ye and Yuan, Haoqi and Liu, Jiazheng and Xu, Chaoyi and Jin, Qin and Lu, Zongqing},
  booktitle={International Conference on Machine Learning},
  year={2026},
  organization={PMLR}
}

@article{zhang2025digflow,
  title={DiG-Flow: Discrepancy-Guided Flow Matching for Robust VLA Models},
  author={Zhang, Wanpeng and Wang, Ye and Luo, Hao and Yuan, Haoqi and Feng, Yicheng and Zheng, Sipeng and Jin, Qin and Lu, Zongqing},
  journal={arXiv preprint arXiv:2512.01715},
  year={2025}
}

@article{chen2025unified,
  title={Unified diffusion vla: Vision-language-action model via joint discrete denoising diffusion process},
  author={Chen, Jiayi and Song, Wenxuan and Ding, Pengxiang and Zhou, Ziyang and Zhao, Han and Tang, Feilong and Wang, Donglin and Li, Haoang},
  journal={arXiv preprint arXiv:2511.01718},
  year={2025}
}

@article{li2025spatial,
  title={Spatial forcing: Implicit spatial representation alignment for vision-language-action model},
  author={Li, Fuhao and Song, Wenxuan and Zhao, Han and Wang, Jingbo and Ding, Pengxiang and Wang, Donglin and Zeng, Long and Li, Haoang},
  journal={arXiv preprint arXiv:2510.12276},
  year={2025}
}

@article{zhang2026world,
  title={Do World Action Models Generalize Better than VLAs? A Robustness Study},
  author={Zhang, Zhanguang and Li, Zhiyuan and Rahmati, Behnam and Yang, Rui Heng and Ma, Yintao and Rasouli, Amir and Pakdamansavoji, Sajjad and Wu, Yangzheng and Zhang, Lingfeng and Cao, Tongtong and others},
  journal={arXiv preprint arXiv:2603.22078},
  year={2026}
}

@inproceedings{song2026reconvla,
  title={Reconvla: Reconstructive vision-language-action model as effective robot perceiver},
  author={Song, Wenxuan and Zhou, Ziyang and Zhao, Han and Chen, Jiayi and Ding, Pengxiang and Yan, Haodong and Huang, Yuxin and Tang, Feilong and Wang, Donglin and Li, Haoang},
  booktitle={Proceedings of the AAAI Conference on Artificial Intelligence},
  volume={40},
  number={22},
  pages={18549--18557},
  year={2026}
}

@inproceedings{wang2026vla,
  title={Vla-adapter: An effective paradigm for tiny-scale vision-language-action model},
  author={Wang, Yihao and Ding, Pengxiang and Li, Lingxiao and Cui, Can and Ge, Zirui and Tong, Xinyang and Song, Wenxuan and Zhao, Han and Zhao, Wei and Hou, Pengxu and others},
  booktitle={Proceedings of the AAAI conference on artificial intelligence},
  volume={40},
  number={22},
  pages={18638--18646},
  year={2026}
}

@article{lee2025molmoact,
  title={Molmoact: Action reasoning models that can reason in space},
  author={Lee, Jason and Duan, Jiafei and Fang, Haoquan and Deng, Yuquan and Liu, Shuo and Li, Boyang and Fang, Bohan and Zhang, Jieyu and Wang, Yi Ru and Lee, Sangho and others},
  journal={arXiv preprint arXiv:2508.07917},
  year={2025}
}

@article{yang2026abot,
  title={ABot-M0: VLA Foundation Model for Robotic Manipulation with Action Manifold Learning},
  author={Yang, Yandan and Zeng, Shuang and Lin, Tong and Chang, Xinyuan and Qi, Dekang and Xiao, Junjin and Liu, Haoyun and Chen, Ronghan and Chen, Yuzhi and Huo, Dongjie and others},
  journal={arXiv preprint arXiv:2602.11236},
  year={2026}
}

@inproceedings{yu2025repa,
  title={Representation Alignment for Generation: Training Diffusion Transformers Is Easier Than You Think},
  author={Sihyun Yu and Sangkyung Kwak and Huiwon Jang and Jongheon Jeong and Jonathan Huang and Jinwoo Shin and Saining Xie},
  year={2025},
  booktitle={International Conference on Learning Representations},
}

@article{wang2025unified,
  title={Unified Vision-Language-Action Model},
  author={Wang, Yuqi and Li, Xinghang and Wang, Wenxuan and Zhang, Junbo and Li, Yingyan and Chen, Yuntao and Wang, Xinlong and Zhang, Zhaoxiang},
  journal={arXiv preprint arXiv:2506.19850},
  year={2025}
}

@misc{nanobananapro2025,
  title={Nano Banana Pro},
  author={{Google DeepMind}},
  howpublished={\url{https://deepmind.google/technologies/gemini/}},
  year={2025},
  note={Built on Gemini~3 Pro. Image generation and editing model.}
}

@article{gabeur2026image,
  title={Image generators are generalist vision learners},
  author={Gabeur, Valentin and Long, Shangbang and Peng, Songyou and Voigtlaender, Paul and Sun, Shuyang and Bao, Yanan and Truong, Karen and Wang, Zhicheng and Zhou, Wenlei and Barron, Jonathan T and others},
  journal={arXiv preprint arXiv:2604.20329},
  year={2026}
}

@article{wang2026diffusion,
  title={Diffusion Model as a Generalist Segmentation Learner},
  author={Wang, Haoxiao and Xiang, Antao and Sun, Haiyang and Sun, Peilin and Pan, Changhao and Chen, Yifu and Hong, Minjie and Wang, Weijie and Chen, Shuang and Chen, Yue and others},
  journal={arXiv preprint arXiv:2604.24575},
  year={2026}
}

@article{fan2026aim,
  title={AIM: Intent-Aware Unified world action Modeling with Spatial Value Maps},
  author={Fan, Liaoyuan and Xu, Zetian and Cao, Chen and Zhang, Wenyao and Yuan, Mingqi and Chen, Jiayu},
  journal={arXiv preprint arXiv:2604.11135},
  year={2026}
}

@article{jeanson2026leveraging,
  title={Leveraging Image Generators to Address Training Data Scarcity: The Gen4Regen Dataset for Forest Regeneration Mapping},
  author={Jeanson, Gabriel and Duclos, David-Alexandre and Larriv{\'e}e-Hardy, William and Cochet, No{\'e} and Boxan, Mat{\v{e}}j and Desch{\^e}nes, Anthony and Pomerleau, Fran{\c{c}}ois and Giguere, Philippe},
  journal={arXiv preprint arXiv:2605.05627},
  year={2026}
}

@article{ye2025imgedit,
  title={Imgedit: A unified image editing dataset and benchmark},
  author={Ye, Yang and He, Xianyi and Li, Zongjian and Lin, Bin and Yuan, Shenghai and Yan, Zhiyuan and Hou, Bohan and Yuan, Li},
  journal={arXiv preprint arXiv:2505.20275},
  year={2025}
}

@misc{glm_image,
  author       = {Zhipu AI},
  title        = {GLM-Image},
  year         = {2026},
  howpublished = {\url{https://huggingface.co/zai-org/GLM-Image}},
}

@article{nextstep,
  title={Nextstep-1: Toward autoregressive image generation with continuous tokens at scale},
  author={Team, NextStep and Han, Chunrui and Li, Guopeng and Wu, Jingwei and Sun, Quan and Cai, Yan and Peng, Yuang and Ge, Zheng and Zhou, Deyu and Tang, Haomiao and others},
  journal={arXiv preprint arXiv:2508.10711},
  year={2025}
}

@article{longcatnext,
  title={LongCat-Next: Lexicalizing Modalities as Discrete Tokens},
  author={Team, Meituan LongCat and Xiao, Bin and Wang, Chao and Li, Chengjiang and Zhang, Chi and Peng, Chong and Yu, Hang and Yang, Hao and Yan, Haonan and Sun, Haoze and others},
  journal={arXiv preprint arXiv:2603.27538},
  year={2026}
}

@article{team2025zimage,
  title={Z-Image: An Efficient Image Generation Foundation Model with Single-Stream Diffusion Transformer},
  author={Z-Image Team},
  journal={arXiv preprint arXiv:2511.22699},
  year={2025}
}

@article{agarwal2026cosmos,
  title={Cosmos 3: Omnimodal World Models for Physical AI},
  author={Agarwal, Niket and Ali, Arslan and Allen, Jon and Antolini, Martin and Aubame, Adeline and Azzolini, Alisson and Bai, Junjie and Bala, Maciej and Balaji, Yogesh and Bapst, Josh and others},
  journal={arXiv preprint arXiv:2606.02800},
  year={2026}
}

@article{wang2026qwen,
  title={Qwen-VLA: Unifying Vision-Language-Action Modeling across Tasks, Environments, and Robot Embodiments},
  author={Wang, Qiuyue and Li, Mingsheng and Guan, Jian and Ye, Jinhui and Xie, Sicheng and Liu, Yitao and Chen, Junhao and Liang, Zhixuan and Zhang, Jie and Hu, Xintong and others},
  journal={arXiv preprint arXiv:2605.30280},
  year={2026}
}

@article{xu2025seeing,
  title={Seeing to Act, Prompting to Specify: A Bayesian Factorization of Vision Language Action Policy},
  author={Xu, Kechun and Zhu, Zhenjie and Chen, Anzhe and Zhao, Shuqi and Huang, Qing and Yang, Yifei and Lu, Haojian and Xiong, Rong and Tomizuka, Masayoshi and Wang, Yue},
  journal={arXiv preprint arXiv:2512.11218},
  year={2025}
}

@article{yu2026maskwam,
  title={MaskWAM: Unifying Mask Prompting and Prediction for World-Action Models},
  author={Yu, Hanyang and Lin, Haitao and Zhang, Jingbo and Zhang, Wenyao and Gu, Chenghao and Li, Heng and Tan, Ping},
  journal={arXiv preprint arXiv:2606.13515},
  year={2026}
}
\bibliographystyle{unsrtnat}

\newpage
\section*{Appendix}

\subsection{Architecture of ImageWAM}
Across the three model variants, namely OmniGen2, FLUX.2[klein], and Ovis-U1, we adopt the MoT structure as our multimodal joint modeling architecture.
\subsubsection{OmniGen2-based ImageWAM}
For the OmniGen2-based ImageWAM variant, we load the LLM component from the corresponding original pretrained Qwen2.5-VL-3B as the LLM backbone, which provides the downstream model with a strong foundation for vision-language alignment. The last-layer hidden states of the Qwen2.5-VL LLM are fed into the OmniGen2 DiT, together with the latent tokens of the reference image and the future noisy frames, for self-attention. In MoT, we extend the original self-attention mechanism into joint self-attention over four types of tokens: language context tokens, visual condition tokens, visual prediction tokens, and action tokens. The visual prediction transformer and the action transformer independently generate their attention QKV representations, which are then concatenated into a complete QKV sequence. The attention mask is configured such that action tokens attend to the other tokens in a one-way manner, while noisy tokens attend only to context tokens, thereby keeping the information in the context tokens clean.

To prevent the visual model from being affected by noisy gradients from the action model during the early stage of training, we adopt an action-head weight-copy initialization strategy similar to \cite{li2026lingbotva,yuan2026fastwam}. Specifically, our Action DiT uses the same architecture as the image editing model. We copy and interpolate the weights of the image editing model to match the size of the Action DiT, and add additional projection layers to support action inputs and outputs. To enable cross-modal attention while maintaining a moderately sized Action DiT, we use a relatively small DiT hidden dimension 1024 while keeping the same attention hidden dimension 2520. The final size of our Action DiT is approximately 760M parameters. %
\subsubsection{FLUX.2-based ImageWAM}
For the FLUX.2-based architecture, the LLM module is the original pretrained Qwen3-4B/8B used by FLUX.2. We similarly extend FLUX.2 into a joint self-attention structure, while modifying the action-head initialization strategy according to the double-stream and single-stream design of FLUX.2. In this setting, the lower layers of the action head are initialized by copying the weights from the image stream in the double-stream stage of FLUX, while the higher layers are initialized from the single-stream blocks of FLUX. The final sizes of the Action DiT in this variant are 642M parameters for the 4B version and 952M parameters for the 9B version. %
\subsubsection{Ovis-U1-based ImageWAM}
For the Ovis-U1-based architecture, we use the Qwen3-1.7B model trained and vision-language fine-tuned by Ovis-U1, and adopt its approximately 1.2B-parameter diffusion visual decoder as our visual editing backbone. In this model, the language context tokens also include vision-language tokens processed by the LLM. Since Ovis-U1 adopts an MMDiT structure similar to FLUX, we use the same Action DiT initialization strategy as in the FLUX.2-based ImageWAM variant. Because this model is relatively small, we do not reduce the DiT hidden dimension. The final size of the Action DiT is 1.1B parameters. %

\subsection{Training Details}

All models are trained on 8 NVIDIA H20 GPUs. Unless otherwise specified, we use DeepSpeed ZeRO-1 for distributed training. For the FLUX.2 9B variant, we use DeepSpeed ZeRO-2 due to its larger model size. All models are trained with bf16 precision and optimized using AdamW. The common training hyperparameters are summarized in Table~\ref{table:training_hyperparams}.

\begin{table}[!htbp]
  \caption{Common training hyperparameters.}
  \label{table:training_hyperparams}
  \begin{center}
    \begin{small}
      \begin{sc}
        \begin{tabular}{ll}
        \toprule
        Parameter & Value \\
        \midrule
        GPUs & 8 NVIDIA H20 \\
        Distributed strategy & DeepSpeed ZeRO-1$^*$ \\
        Precision & bf16 \\
        Optimizer & AdamW \\
        Optimizer betas & $(0.9, 0.95)$ \\
        Learning rate & $1 \times 10^{-4}$ \\
        Weight decay & $1 \times 10^{-2}$ \\
        LR scheduler & Warmup cosine \\
        Warmup steps & $0.05 T_{\mathrm{total}}$ \\
        Minimum LR & $0.01 \times \mathrm{lr}$ \\
        Gradient clipping & 1.0 \\
        \bottomrule
        \end{tabular}
      \end{sc}
    \end{small}
    
    \begin{small}
    $^*$For FLUX.2 9B, we use ZeRO-2 for VRAM compatibility.
    \end{small}
  \end{center}
  \vskip -0.1in
\end{table}

On LIBERO, we horizontally concatenate the two camera views and resize the resulting image to $224 \times 448$. The model predicts the future observation 16 frames ahead, together with an action chunk of length 16. We train on the merged dataset of the four LIBERO suites for 10 epochs.

On RoboTwin, we first resize the two wrist-view images to a smaller resolution and horizontally concatenate them. The concatenated wrist views are then vertically concatenated with the main-view image, and the final input is resized to $288 \times 256$. The model also predicts the future observation 16 frames ahead and an action chunk of length 16. We train the models for 5 epochs.

On Real-World Dataset, we follow the same preprocess in RoboTwin, predicting 16 action steps and training on all four task for 10 epoch.

\begin{table}[!htbp]
  \caption{Dataset-specific training configurations.}
  \label{table:dataset_training_config}
  \begin{center}
    \begin{small}
      \begin{sc}
        \begin{tabular}{lll}
        \toprule
        Parameter & LIBERO & RoboTwin \\
        \midrule
        Input views & 2 views & 3 views \\
        View layout & Horizontal & Wrist-horizontal + vertical \\
        Input resolution & $224 \times 448$ & $288 \times 256$ \\
        Future horizon & 16 frames & 16 frames \\
        Action chunk length & 16 & 16 \\
        Training epochs & 10 & 5 \\
        \bottomrule
        \end{tabular}
      \end{sc}
    \end{small}
  \end{center}
  \vskip -0.1in
\end{table}

\begin{table}[!htbp]
  \caption{Training cost and batch size.}
  \label{table:training_cost}
  \begin{center}
    \begin{small}
      \begin{sc}
        \begin{tabular}{llll}
        \toprule
        Benchmark & Model & Time & Batch/GPU \\
        \midrule
        LIBERO & OmniGen2 & 18 hours & 12 \\
        LIBERO & Ovis-U1 & 18 hours & 16 \\
        LIBERO & FLUX.2 4B & 18 hours & 10 \\
        LIBERO & FLUX.2 9B & 1.6 days & 12 \\
        \midrule
        RoboTwin & OmniGen2 & 5 days & 48$^\dagger$ \\
        RoboTwin & FLUX.2 4B & 5 days & 48$^\dagger$ \\
        \midrule
        Real-World Robot & OmniGen2 & 18 hours & 16 \\
        \bottomrule
        \end{tabular}
      \end{sc}
    \end{small}
    
  \begin{small}
  $^\dagger$ Effective per-GPU batch size with gradient accumulation over three steps.
  \end{small}
  \end{center}
  
\end{table}
\section{Efficiency Optimization}
To further optimize inference latency, we also evaluate on our model the prefix-only attention training and image-denoising-free inference strategy, similar to that adopted in FastWAM. In addition, we explore model optimization with `torch.compile` and static CUDA graphs. The inference latency results are reported in Table~\ref{table:inference_latency}, where all models use three action denoising steps during inference. We observe that adding compilation brings nearly a 3$\times$ overall speedup, mainly due to the improved efficiency of the action head. This is because, under typical action chunk lengths, the number of action tokens is relatively small, making the parallel efficiency of the Action DiT often suboptimal.

\begin{table}[!htbp]
  \caption{Inference latency and relative speedup. Speedup is computed with respect to FastWAM with one video denoising step.}
  \label{table:inference_latency}
  \begin{center}
    \begin{small}
      \begin{sc}
        \begin{tabular}{lcc}
        \toprule
        Variant & Latency (ms) & Speedup \\
        \midrule
        FastWAM (1$\times$ Vid. Denoise) & 302 & $1.00\times$ \\
        ImageWAM (1$\times$ Vid. Denoise) & 263 & $1.15\times$ \\
        FastWAM (Prefix Only) & 194 & $1.56\times$ \\
        \quad + Compiled & 80 & $3.78\times$ \\
        ImageWAM (Prefix Only) & 198 & $1.53\times$ \\
        \quad + Action Loop Compile & 85 & $3.55\times$ \\
        \quad + Image Prefill Compile & 77 & $3.92\times$ \\
        \quad + Action Static Graph & 69 & $4.38\times$ \\
        \bottomrule
        \end{tabular}
      \end{sc}
    \end{small}
  \end{center}
  \vskip -0.1in
\end{table}
\section{Real-World Experiments Detail}

\subsection{Task settings and evaluation in Real-world Tasks}
\paragraph{Task Settings.} 
To evaluate the capability and generalizability of ImageWAM, we design four representative and challenging real-world manipulation tasks, including:
(1) \textbf{Stack Three Bowls (T1)}, stacking three green nested bowls; 
(2) \textbf{Fold Towel (T2)}, folding a fabric towel; 
(3) \textbf{Open Drawer \& Store Marker (T3)}, which involves opening a drawer, placing a marker inside, and closing the drawer; and 
(4) \textbf{Hang Cup On Rack (T4)}, hanging a mug onto a designated peg on a wooden stand. 
We collect an average of 100 demonstrations per task. Each model is evaluated over 50 trials per task. The execution success rate is reported as the primary performance metric.

\section{RoboTwin Evaluation Results}

Here we present the per-task results on RoboTwin evaluation in Table~\ref{tab:robotwin_detail}.

\begin{table*}[t]
\centering
\tiny
\caption{Per-task success rates on RoboTwin under clean and randomized evaluation settings.}
\label{tab:robotwin_detail}
\setlength{\tabcolsep}{2pt}
\resizebox{\textwidth}{!}{%
\begin{tabular}{lcc|cc|cc|cc|cc|cc|cc}
\toprule
Task & \multicolumn{2}{c}{\makecell{\textbf{\method{}}\\\textbf{Flux.2 4B (Ours)}}} & \multicolumn{2}{c}{\makecell{\textbf{\method{}}\\\textbf{OmniGen2 (Ours)}\\(50 trials)}} & \multicolumn{2}{c}{Fast-WAM-IDM} & \multicolumn{2}{c}{\makecell{Fast-WAM\\w.o. co-train}} & \multicolumn{2}{c}{LingBot-VA} & \multicolumn{2}{c}{$\pi_{0.5}$} & \multicolumn{2}{c}{Motus} \\
\cmidrule(lr){2-3}\cmidrule(lr){4-5}\cmidrule(lr){6-7}\cmidrule(lr){8-9}\cmidrule(lr){10-11}\cmidrule(lr){12-13}\cmidrule(lr){14-15}
 & Clean & Rand. & Clean & Rand. & Clean & Rand. & Clean & Rand. & Clean & Rand. & Clean & Rand. & Clean & Rand. \\
\midrule
Adjust Bottle & \textbf{100} & 99 & \textbf{100} & \textbf{100} & 94 & 99 & 98 & \textbf{100} & 90 & 94 & \textbf{100} & 99 & 89 & 93 \\
Beat Block Hammer & 98 & \textbf{99} & \textbf{100} & 98 & 98 & 98 & 80 & 92 & 96 & 98 & 96 & 93 & 95 & 88 \\
Blocks Ranking RGB & 96 & \textbf{99} & \textbf{100} & 96 & \textbf{100} & \textbf{99} & 88 & 86 & 99 & 98 & 92 & 85 & 99 & 97 \\
Blocks Ranking Size & \textbf{96} & \textbf{100} & 86 & 92 & 79 & 90 & 56 & 62 & 94 & 96 & 49 & 26 & 75 & 63 \\
Click Alarmclock & 98 & \textbf{100} & \textbf{100} & \textbf{100} & 98 & \textbf{100} & \textbf{100} & 98 & 99 & \textbf{100} & 98 & 89 & \textbf{100} & \textbf{100} \\
Click Bell & \textbf{100} & \textbf{100} & \textbf{100} & \textbf{100} & \textbf{100} & 96 & \textbf{100} & \textbf{100} & \textbf{100} & \textbf{100} & 99 & 66 & \textbf{100} & \textbf{100} \\
Dump Bin Bigbin & \textbf{96} & 90 & 92 & 88 & 93 & \textbf{98} & 92 & 94 & 89 & 96 & 92 & 97 & 95 & 91 \\
Grab Roller & \textbf{100} & \textbf{100} & \textbf{100} & \textbf{100} & \textbf{100} & \textbf{100} & \textbf{100} & \textbf{100} & \textbf{100} & \textbf{100} & \textbf{100} & \textbf{100} & \textbf{100} & \textbf{100} \\
Handover Block & 96 & \textbf{95} & 94 & 84 & 97 & 94 & 58 & 46 & \textbf{99} & 78 & 66 & 57 & 86 & 73 \\
Handover Mic & \textbf{100} & \textbf{100} & \textbf{100} & \textbf{100} & 98 & 99 & \textbf{100} & \textbf{100} & 94 & 96 & 98 & 97 & 78 & 63 \\
Hanging Mug & \textbf{74} & \textbf{84} & 50 & 56 & 66 & 62 & 28 & 40 & 40 & 28 & 18 & 17 & 38 & 38 \\
Lift Pot & \textbf{100} & \textbf{100} & \textbf{100} & \textbf{100} & \textbf{100} & \textbf{100} & 92 & 90 & \textbf{100} & 99 & 96 & 85 & 96 & 99 \\
Move Can Pot & 96 & 98 & 96 & 92 & \textbf{97} & \textbf{100} & 80 & 68 & 94 & 97 & 51 & 55 & 34 & 74 \\
Move Pillbottle Pad & 98 & \textbf{100} & 98 & 98 & 98 & \textbf{100} & 88 & 96 & \textbf{99} & 99 & 84 & 61 & 93 & 96 \\
Move Playingcard Away & \textbf{100} & 99 & \textbf{100} & \textbf{100} & 99 & \textbf{100} & 94 & 96 & \textbf{100} & 99 & 96 & 84 & \textbf{100} & 96 \\
Move Stapler Pad & 67 & 60 & 74 & 82 & 89 & \textbf{85} & 64 & 78 & \textbf{91} & 79 & 56 & 42 & 83 & \textbf{85} \\
Open Laptop & 98 & 98 & 96 & \textbf{100} & 92 & 92 & \textbf{100} & 98 & 92 & 94 & 90 & 96 & 95 & 91 \\
Open Microwave & 97 & \textbf{94} & \textbf{98} & 82 & 54 & 53 & 46 & 52 & 82 & 86 & 34 & 77 & 95 & 91 \\
Pick Diverse Bottles & 84 & 88 & 84 & \textbf{92} & 87 & 89 & 58 & 62 & 89 & 82 & 81 & 71 & \textbf{90} & 91 \\
Pick Dual Bottles & 96 & 98 & \textbf{100} & \textbf{100} & \textbf{100} & 98 & 80 & 74 & \textbf{100} & 99 & 93 & 63 & 96 & 90 \\
Place A2B Left & 95 & 93 & 94 & \textbf{100} & \textbf{97} & 96 & 84 & 92 & \textbf{97} & 93 & 87 & 82 & 88 & 79 \\
Place A2B Right & 96 & 94 & 96 & \textbf{98} & 94 & \textbf{98} & 88 & 84 & \textbf{97} & 95 & 87 & 84 & 91 & 87 \\
Place Bread Basket & 96 & 92 & 90 & 94 & 91 & \textbf{97} & 74 & 76 & \textbf{97} & 95 & 77 & 64 & 91 & 94 \\
Place Bread Skillet & 90 & 89 & 92 & 90 & 90 & \textbf{95} & \textbf{98} & 84 & 95 & 90 & 85 & 66 & 86 & 83 \\
Place Burger Fries & 95 & \textbf{100} & \textbf{100} & \textbf{100} & 97 & 99 & 94 & 96 & 97 & 95 & 94 & 87 & 98 & 98 \\
Place Can Basket & 74 & 72 & \textbf{82} & 76 & 37 & 28 & 72 & 72 & 81 & \textbf{84} & 62 & 62 & 81 & 76 \\
Place Cans Plasticbox & 99 & 97 & \textbf{100} & 94 & 98 & 96 & 98 & 96 & \textbf{100} & \textbf{99} & 94 & 84 & 98 & 94 \\
Place Container Plate & 98 & 98 & 98 & 98 & \textbf{100} & 96 & 94 & 98 & 99 & 97 & 99 & 95 & 98 & \textbf{99} \\
Place Dual Shoes & 81 & 81 & 90 & 84 & 85 & 87 & 80 & 74 & \textbf{94} & \textbf{89} & 75 & 75 & 93 & 87 \\
Place Empty Cup & \textbf{100} & \textbf{100} & \textbf{100} & \textbf{100} & \textbf{100} & \textbf{100} & \textbf{100} & \textbf{100} & \textbf{100} & \textbf{100} & \textbf{100} & 99 & 99 & 98 \\
Place Fan & 95 & 94 & 94 & 88 & 97 & \textbf{95} & 80 & 88 & \textbf{99} & 93 & 87 & 85 & 91 & 87 \\
Place Mouse Pad & 84 & 93 & 90 & 84 & \textbf{97} & 93 & 64 & 76 & 93 & \textbf{96} & 60 & 39 & 66 & 68 \\
Place Object Basket & 86 & 83 & \textbf{92} & \textbf{90} & 87 & 82 & 82 & \textbf{90} & 91 & 88 & 80 & 76 & 81 & 87 \\
Place Object Scale & 97 & 96 & 92 & 98 & \textbf{99} & \textbf{99} & 86 & 80 & 96 & 95 & 86 & 80 & 88 & 85 \\
Place Object Stand & 98 & 98 & \textbf{100} & 92 & 96 & \textbf{100} & 82 & 92 & 99 & 96 & 91 & 85 & 98 & 97 \\
Place Phone Stand & \textbf{100} & \textbf{100} & 98 & 98 & 99 & 99 & 90 & 92 & 97 & 97 & 81 & 81 & 87 & 86 \\
Place Shoe & 97 & 95 & 94 & 96 & 95 & \textbf{98} & 92 & 90 & 98 & \textbf{98} & 92 & 93 & \textbf{99} & 97 \\
Press Stapler & \textbf{97} & \textbf{100} & 90 & 94 & 50 & 57 & 80 & 80 & 85 & 82 & 87 & 83 & 93 & 98 \\
Put Bottles Dustbin & \textbf{97} & 91 & 92 & \textbf{96} & \textbf{97} & 92 & 78 & 88 & 87 & 91 & 84 & 79 & 81 & 79 \\
Put Object Cabinet & 91 & 89 & 90 & \textbf{96} & \textbf{93} & 90 & 88 & 84 & 85 & 87 & 80 & 79 & 88 & 71 \\
Rotate QRcode & 87 & \textbf{92} & 82 & 90 & 91 & 86 & 82 & 78 & \textbf{96} & 91 & 89 & 87 & 89 & 73 \\
Scan Object & 94 & 90 & 94 & 86 & 93 & 90 & 78 & 86 & \textbf{96} & \textbf{91} & 72 & 65 & 67 & 66 \\
Shake Bottle & \textbf{100} & \textbf{100} & \textbf{100} & \textbf{100} & \textbf{100} & \textbf{100} & \textbf{100} & \textbf{100} & \textbf{100} & 97 & 99 & 97 & \textbf{100} & 97 \\
Shake Bottle Horizontally & \textbf{100} & \textbf{100} & \textbf{100} & \textbf{100} & \textbf{100} & \textbf{100} & \textbf{100} & \textbf{100} & \textbf{100} & 99 & 99 & 99 & \textbf{100} & 98 \\
Stack Blocks Three & 96 & 97 & \textbf{100} & \textbf{100} & 99 & 95 & 90 & 90 & 99 & 98 & 91 & 76 & 91 & 95 \\
Stack Blocks Two & 99 & \textbf{100} & \textbf{100} & \textbf{100} & \textbf{100} & \textbf{100} & \textbf{100} & 98 & \textbf{100} & 98 & 97 & \textbf{100} & \textbf{100} & 98 \\
Stack Bowls Three & 78 & 83 & 84 & 86 & 85 & 83 & 66 & 82 & \textbf{86} & 83 & 77 & 71 & 79 & \textbf{87} \\
Stack Bowls Two & 94 & 97 & 92 & \textbf{98} & 94 & 96 & 90 & \textbf{98} & 94 & \textbf{98} & 95 & 96 & \textbf{98} & \textbf{98} \\
Stamp Seal & 79 & 84 & 76 & 84 & \textbf{99} & 94 & 60 & 78 & 96 & \textbf{97} & 79 & 55 & 93 & 92 \\
Turn Switch & 77 & \textbf{79} & 54 & 70 & 59 & 74 & 66 & 66 & 44 & 45 & 62 & 54 & \textbf{84} & 78 \\
\midrule
\textbf{Average} & \textbf{93.20} & \textbf{93.56} & 92.48 & 92.80 & 91.16 & 91.34 & 82.76 & 84.80 & 92.90 & 91.50 & 82.74 & 76.76 & 88.66 & 87.02 \\
\bottomrule
\end{tabular}
}

\end{table*}

\end{document}